\documentclass{article}

\PassOptionsToPackage{numbers, compress}{natbib}
\usepackage[utf8]{inputenc}
\usepackage{newunicodechar}
\newunicodechar{，}{,}
\usepackage[main, final]{neurips_2026}

\usepackage[utf8]{inputenc}
\usepackage[T1]{fontenc}
\usepackage[colorlinks=true, linkcolor=black, citecolor=blue!80, urlcolor=black]{hyperref}
\usepackage{url}
\usepackage{booktabs}
\usepackage{amsfonts}
\usepackage{nicefrac}
\usepackage{microtype}
\usepackage{xcolor}
\usepackage{amsmath}
\usepackage{enumitem}
\usepackage{graphicx}
\usepackage{booktabs}
\usepackage{multirow}
\usepackage[table]{xcolor}
\definecolor{lightgray}{gray}{0.85}
\definecolor{lightblue}{RGB}{220, 235, 255}

\usepackage{xcolor}
\usepackage{amssymb}
\usepackage{wrapfig}
\newcommand{\reddrop}[1]{$_{\textcolor{red}{\downarrow\text{\!#1}}}$}
\usepackage{algorithm}
\usepackage{algpseudocode}
\usepackage{amsmath}
\usepackage{xcolor} 
\usepackage{float}

\title{Learn from the Gap: Differential-Aware Advantage Pruning with Adaptive Rollout Sampling for GRPO} 

\author{
\setlength{\tabcolsep}{2pt}
\begin{tabular}{ccccc}
Jiahua Yang$^1$ &
Zhiwei Yang$^{1,*}$ &
Xianpeng Zhang$^2$ &
Dongyu Chen$^2$ & Xing Chen$^3$\\[4pt]
Tianhuang Su$^2$ &
Haonan Lu$^2$ &
Quanlong Guan$^1$ &
Kai Tang$^2$ &
Chuangchuang Wang$^{2,*}$
\end{tabular}
\\[10pt]
$^1$Guangdong Institute of Smart Education, Jinan University, Guangzhou, China\\
$^2$OPPO AI Center, Shenzhen, China\\
$^3$Ragentile Intelligence Inc, Edmonton, Canada \\[4pt]
yangjiahua@stu2024.jnu.edu.cn,\ 
yangzw@jnu.edu.cn,\ 
wangchuangchuang@oppo.com
}

\begin{document}

\maketitle

\begingroup
\renewcommand{\thefootnote}{\fnsymbol{footnote}}
\footnotetext[1]{Corresponding authors: Zhiwei Yang and Chuangchuang Wang.}
\endgroup

\begin{abstract}
Recently, Group Relative Policy Optimization (GRPO) and its variants have been developed for policy optimization and demonstrated notable performance gains.  However, these methods usually incur substantial computational overhead due to per-question multi-rollout sampling and repeated per-token probability evaluation across rollouts. Furthermore, low-information or highly homogeneous trajectories can degrade downstream learning signal efficiency, hindering model optimization and limiting final performance. 
To address these issues, we propose FastRL, a novel plug-and-play reinforcement learning framework that simultaneously improves training efficiency and the effectiveness of policy learning. Specifically, 1) We introduce an advantage-aware pruning strategy to selectively preserve high-advantage trajectories while maximizing inter-trajectory gradient diversity. 2) Then, we design an adaptive rollout sampling mechanism to dynamically adjust the sampling scale across different training stages based on historical pruning distributions, balancing exploration adequacy and computational efficiency. 
Experiments demonstrate that FastRL can be seamlessly integrated into GRPO, DAPO, and GSPO variants, achieving an average 2.07$\times$ training speedup on Geometry3K and GeoQA8K-R1V, along with an approximately 1.64\% improvement in average accuracy on visual reasoning benchmarks. Source codes will be available at https://github.com/Nicozwy/FastRL. 

\end{abstract}

\section{Introduction}
\begin{figure*}[t]
\centering
\includegraphics[width=1\textwidth]{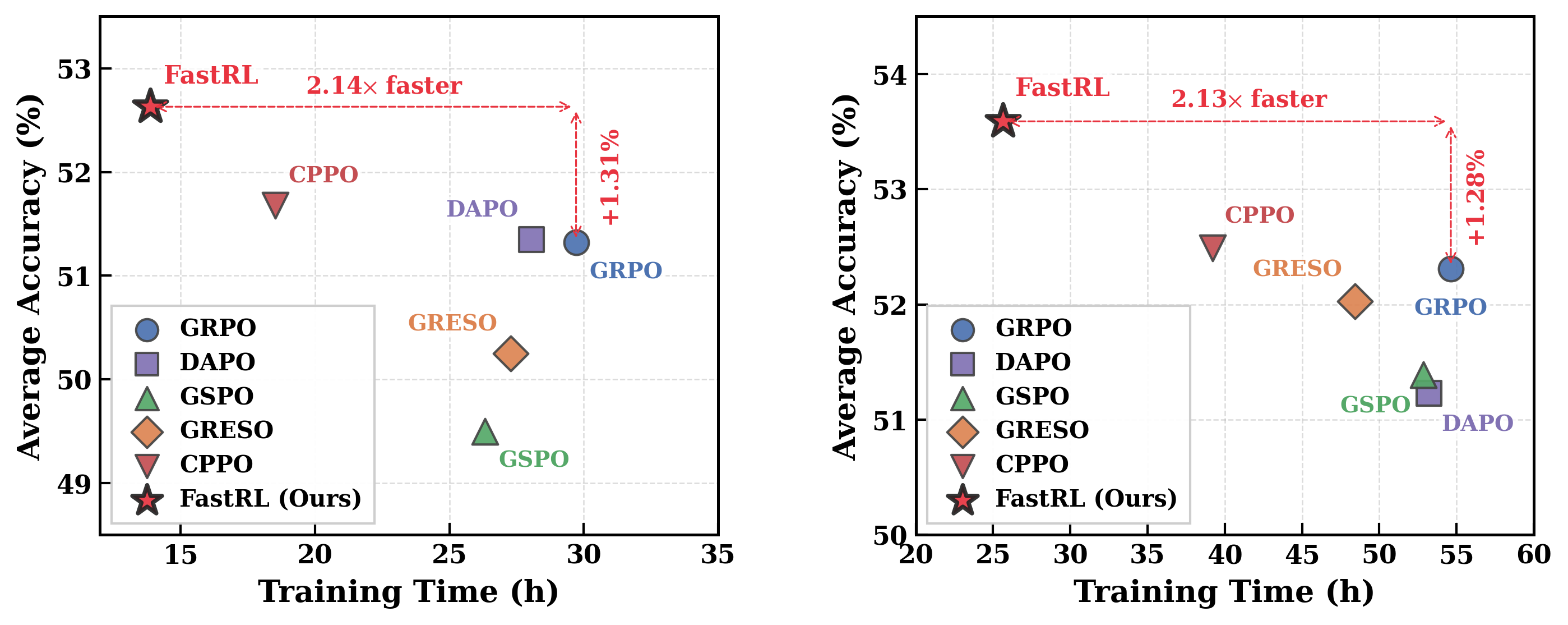} 
\caption{Comparison of training efficiency and out-of-domain performance across reinforcement learning methods using Qwen2.5-VL-7B-Instruct on Geometry3K (left) and GeoQA8K-R1V (right).
}
\label{fig:benchmark_comparison}
\end{figure*}

Reinforcement Learning (RL) optimizes model generation policies through interactive reward signals and sequential optimization~\cite{shao2024deepseekmath,wang2025_1000}, which has proven effective in eliciting complex reasoning for mathematics, coding, and scientific reasoning tasks. Despite the strong reasoning performance of policy gradient methods, they incur heavy computational costs and memory overhead.  
To mitigate the excessive memory overhead inherent in Proximal Policy Optimization (PPO)~\cite{schulman2017proximal}, 
Group Relative Policy Optimization (GRPO)~\cite{shao2024deepseekmath} is proposed as a lightweight alternative. It eliminates the standalone critic network and computes advantages via group-wise relative rewards, thereby drastically cutting memory consumption. 
Subsequently, a series of GRPO variants, such as Dynamic Sampling Policy Optimization (DAPO)~\cite{yu2025dapo} and Group Sequence Policy Optimization (GSPO)~\cite{zheng2025group}, have been successively developed. Nevertheless, GRPO necessitates sampling multiple completion trajectories for each prompt and performing repeated per-token evaluation across them, which still imposes non-negligible computational overhead. 
Even worse, since not all sampled trajectories contribute equally to policy updating ~\cite{lin2025cppo}, low-information and highly homogeneous trajectories tend to degrade learning signals, thereby hindering policy optimization and constraining overall performance. 

Recently, a series of improved variants of the GRPO paradigm have been developed, targeting further advances in computational efficiency and training stability.  For example, GRPO with Efficient Selective Rollout (GRESO)~\cite{zheng2025act} imposes input-level regularization by adopting a curriculum learning strategy that excludes overly simple questions and excessively hard instances, thereby alleviating the full training burden. Instead, Completion Pruning Policy Optimization (CPPO)~\cite{lin2025cppo} reduces computational redundancy at the output level by globally pruning trajectory-wise advantage values via a fixed threshold throughout training. While these methods yield substantial computational overhead reduction, considerable redundant overhead remains and constrains further training speedup. As shown in Figure~\ref{fig:benchmark_comparison}, although DAPO, GSPO and GRESO can accelerate the training of various models to some extent, they usually yield inferior final performance. While CPPO substantially boosts training efficiency, it only achieves marginal gains in downstream performance (By contrast, our FastRL outperforms all baselines, achieving up to 2.14× training speedup). 
Overall, existing GRPO variants inherently suffer from a prevalent efficiency-performance trade-off, failing to maintain high task accuracy while pursuing faster training. 

To this end, we present FastRL, an efficient GRPO-style training framework that simultaneously boosts training efficiency and policy learning performance. Built on the core insight of learning from informative differences, FastRL integrates two complementary components to streamline training while preserving performance: 1) Maximizing differential advantage pruning filters low-information or homogeneous trajectories by leveraging the distribution of advantage values within each group, retaining only trajectories with large differential signals and high gradient contributions to eliminate unnecessary computation. 2) Adaptive rollout sampling dynamically adjusts the sampling scale throughout training, utilizing historical pruning statistics to balance computational overhead and sample effectiveness.  Thus, these two mechanisms work collaboratively to train without compromising the stability and quality of policy learning. 
Our contributions are summarized as follows:
\begin{itemize}[leftmargin=*]
 \setlength{\itemsep}{0pt}  
\item We propose FastRL, a unified plug-and-play framework for GRPO variants. It resolves core efficiency and performance bottlenecks of existing methods, boosting training speed and policy learning while enabling flexible deployment across diverse GRPO-family algorithms. 

\item We design a differential-aware advantage pruning strategy and an adaptive dynamic sampling mechanism to alleviate computational bottlenecks in GRPO variants. The two modules respectively select discriminative trajectories based on intra-group advantage distribution to reduce homogeneous trajectories and dynamically regulate rollout scale using historical pruning statistics. 

\item Experimental results show that FastRL achieves up to 2.14$\times$ end-to-end training speedup on visual reasoning benchmarks while outperforming leading GRPO variants, highlighting its efficacy in enhancing the efficiency of reinforcement learning without sacrificing performance gains. 
\end{itemize} 

\section{Related Work}
\textbf{Reinforcement Learning.} 
Reinforcement learning (RL) has emerged as a dominant paradigm for empowering large models and has achieved remarkable success across complex real-world tasks, spanning diverse domains including game play~\cite{ye2021mastering,schrittwieser2020mastering}, autonomous driving~\cite{wu2022uncertainty,fang2022offline},
coding~\cite{le2022coderl,sakharova2025integrating}, and media content generation~\cite{zhou2025dreamdpo,lopez2020deep}.
A representative example is PPO~\cite{schulman2017proximal}, which stabilizes policy optimization across diverse tasks via clipped surrogate objectives. However, its reliance on a dedicated critic network for advantage estimation introduces substantial computational and memory overhead. To address this limitation, GRPO~\cite{shao2024deepseekmath} eliminates the critic network by estimating baselines through group-wise relative rewards, offering a more efficient alternative. Building on this, a series of GRPO-based methods have been proposed, including DAPO~\cite{yu2025dapo} and GSPO~\cite{zheng2025group}. 
Despite their promise, these methods still face several key challenges:  1) Repeated rollout sampling of trajectories, followed by gradient computation and backpropagation, incurs substantial time and computational cost, making training efficiency a central bottleneck in practical deployment.
2) Since multiple trajectories for the same instance are generated by the same policy model, high homogeneity is nearly unavoidable. This limits the informativeness of the learning signal and introduces reasoning bias, ultimately inducing mode collapse in the policy. As the policy entropy continuously decreases toward zero~\cite{lu2026prompt}, outputs become increasingly deterministic and training may eventually collapse. 

\textbf{GRPO-based Training Acceleration.} 
The computational cost of GRPO-style methods broadly stems from two distinct stages: the rollout stage and the gradient computation and policy update stage ~\cite{wang2025_1000,su2025reinforcement}.
During the rollout stage, recent studies~\cite{zhang2026fastgrpo,kim2026mc,li2025limr,yang2026srjudge} have shown that only a small subset of the original training data is sufficient to improve the model's reasoning capabilities. Motivated by this observation, prior work has introduced curriculum learning strategies into the RL training process. For instance, ~\citet{zheng2025act} propose to filter training questions based on their historical difficulty and reward variance, thereby reducing the number of questions involved in training. However, such approaches typically require frequent recording and access to question-level statistics, incurring additional storage and computational overhead, which limits their overall efficiency gains.
In the gradient computation and update stage, ~\citet{lin2025cppo} propose a unified pruning strategy based on the absolute advantage, where trajectories with low absolute advantage are discarded to reduce both forward and backward computation costs. While effective, this strategy applies a uniform criterion across all questions and is therefore relatively coarse-grained. As a result, it may suffer from a “bucket effect”~\cite{lin2025cppo}, where questions that require more aggressive pruning are insufficiently filtered, while others are over-pruned. Furthermore, since this method focuses primarily on gradient computation, it cannot alleviate the latency incurred in the rollout phase. 

\section{Method}

\begin{figure*}[t]
\centering
\includegraphics[width=1\textwidth]{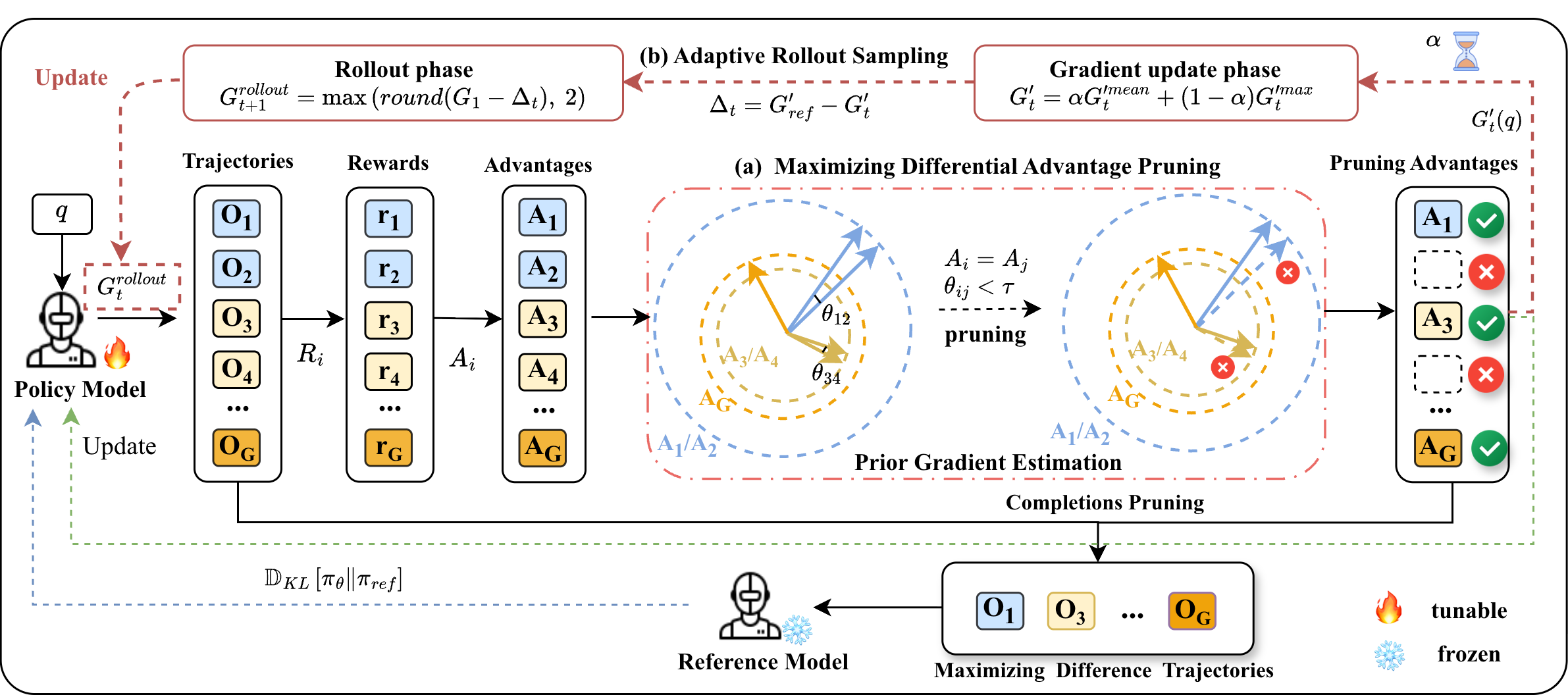}
\caption{The proposed FastRL framework. (a) Maximizing Differential Advantage Pruning (MDAP) identifies and prunes uninformative trajectories (i.e., $\theta_{ij} <\tau $) and homogeneous trajectories (i.e., $A_i = A_j$); (b) Adaptive Rollout Sampling (ARS) dynamically adjusts the rollout budget for each iteration based on the accumulated pruning distribution.  
}
\label{framework}
\end{figure*}

\subsection{Preliminaries}
\label{sec:Preliminaries}
The core idea of GRPO is to construct the advantage function using relative rewards within a group (e.g., mean-centered or normalized rewards), which effectively introduces an implicit baseline to guide policy updates.
Specifically, for each question $q$ sampled from the data distribution $P(Q)$, GRPO uses the old policy $\pi_{\theta_{\text{old}}}$ to generate $G$ candidate outputs, denoted as $O(q) = \{o_1, o_2, \dots, o_G\}$. The policy $\pi_\theta$ is then optimized by maximizing the following objective:
\begin{equation}
\begin{aligned}
\mathcal{J}_{\mathrm{GRPO}}(\theta)
= \mathbb{E}_{q \sim P(Q), \{o_i\}_{i=1}^G \sim \pi_{\theta_{\mathrm{old}}}(o|q)}
\Big\{
\frac{1}{G} \sum_{i=1}^{G} \frac{1}{|o_i|} \sum_{t=1}^{|o_i|}
\Big\{
\min \Big[
\frac{\pi_\theta(o_{i,t}\!\mid\!q,o_{i,<t})}{\pi_{\theta_{\mathrm{old}}}(o_{i,t}\!\mid\!q,o_{i,<t})} A_i,
\\
\qquad\qquad
\mathrm{clip}\!\Big(
\frac{\pi_\theta(o_{i,t}\!\mid\!q,o_{i,<t})}{\pi_{\theta_{\mathrm{old}}}(o_{i,t}\!\mid\!q,o_{i,<t})},
\,1-\epsilon,\,1+\epsilon
\Big) A_i
\Big]
- \beta D_{\mathrm{KL}}\!\left[\pi_\theta \parallel \pi_{\mathrm{ref}}\right]
\Big\}
\Big\}.
\end{aligned}
\end{equation}

where $\pi_{\text{ref}}$ denote the reference model. The clipping coefficient $\epsilon$ is used to constrain the magnitude of policy updates, while $\beta$ is the regularization coefficient that controls the weight of the Kullback-Leibler (KL) divergence penalty. 
The advantage $A_i$ is computed from the reward set ${r_1, r_2, \dots, r_G}$ corresponding to sampled outputs within the same group, as defined in Eq.~\ref{eq:advantage}.
Furthermore, to quantify the contribution of each rollout trajectory to the training gradient, we derive the gradient of the GRPO objective, denoted as
$\nabla_\theta \mathcal{J}_{\text{GRPO}}^{\text{clip}}(\theta)$.
Since the $\min$ and $\mathrm{clip}$ operators are piecewise linear in the importance ratio, differentiating the surrogate activates only one of its two branches, and the gradient with respect to the ratio vanishes whenever the clipped branch is selected. Accordingly, we expand it as follows:
\begin{align}
& \nabla_\theta \mathcal{J}_{\text{GRPO}}^{\text{clip}}(\theta)
= \mathbb{E}_{\substack{q \sim P(Q) \\ \{o_i\}_{i=1}^G \sim \pi_{\theta_{\text{old}}}(O|q)}}
\Big\{ \frac{1}{G} \sum_{i=1}^G \frac{1}{|o_i|} \sum_{t=1}^{|o_i|}
\Big[
c_{i,t}\,
\frac{\pi_\theta(o_{i,t}\mid q,o_{i,<t})}{\pi_{\theta_{\text{old}}}(o_{i,t}\mid q,o_{i,<t})} A_i
\notag \\
&\qquad\qquad
+ \beta \Big(
\frac{\pi_{\text{ref}}(o_{i,t}\mid q,o_{i,<t})}{\pi_\theta(o_{i,t}\mid q,o_{i,<t})} - 1
\Big)
\Big] \nabla_\theta \log \pi_\theta(o_{i,t}\mid q,o_{i,<t})
\Big\}
\end{align}
Therefore, the gradient for each rollout trajectory is:
\begin{equation}
\begin{aligned}
g_i =& \frac{1}{|o_i|}\sum_{t=1}^{|o_i|}
\Big[
c_{i,t}\,r_{i,t}A_i
+ \beta\Big(
\frac{\pi_\text{ref}(o_{i,t}\mid q,o_{i,<t})}{\pi_\theta(o_{i,t}\mid q,o_{i,<t})} - 1
\Big)
\Big]\quad \nabla_\theta\log\pi_\theta(o_{i,t}\mid q,o_{i,<t})
\end{aligned}
\label{eq:grad}
\end{equation}
where $r_{i,t}$ denotes the importance sampling ratio, which corrects for the distribution mismatch between the current policy $\pi_\theta$ and the behavior policy $\pi_{\theta_{\text{old}}}$. The binary coefficient $c_{i,t}=\mathbf{1}[A_i\!\geq\!0]\,\mathbf{1}[r_{i,t}\!\leq\!1\!+\!\epsilon]+\mathbf{1}[A_i\!<\!0]\,\mathbf{1}[r_{i,t}\!\geq\!1\!-\!\epsilon]$ is the clipping indicator induced by the $\min$ and $\mathrm{clip}$ operators, which switches off the gradient of a token once its ratio leaves the trust region along the direction favored by the advantage.
The term $\nabla_\theta \log \pi_\theta(o_{i,t} \mid q, o_{i,<t})$ represents the policy gradient direction, which determines the update direction of model parameters, while the update magnitude is jointly modulated by the advantage, the importance ratio, and the clipping indicator. 
In addition, the last term penalizes the deviation of the current policy from the reference policy. However, due to clipping, many recent works have begun omitting the KL divergence regularization term in practice.

However, in practice, GRPO-based methods that utilize all sampled trajectories often introduce a large amount of low-information or highly homogeneous gradient signals. As implied by Eq.~\ref{eq:grad}, at the individual level, a trajectory whose advantage approaches zero contributes negligibly to the gradient. At the group level, structurally similar trajectories yield highly correlated gradient signals, implicitly overweighting frequent reasoning patterns and biasing policy updates toward dominant modes at the expense of diverse exploration. Moreover, since trajectories for the same input are generated by the same policy, a high degree of homogeneity is nearly unavoidable, leading to rapidly diminishing marginal information gain. Thus, filtering homogeneous trajectories and retaining diverse, representative ones is essential for improving gradient signal quality and achieving a better trade-off between training efficiency and model performance.

\subsection{Maximizing Differential Advantage Pruning}

As illustrated in Figure~\ref{framework}, for a given question $q$, we denote its $G$ rollout trajectories as:
$O(q)=\{o_1,o_2,\dots,o_G\}.$
Each trajectory is associated with a reward signal reflecting both output quality and structural compliance. Specifically, the reward $r_i$ for trajectory $o_i$ is defined as follows:
\begin{equation}
r_i = R_{\text{format}}(o_i) + R_{\text{accuracy}}(o_i).
\end{equation}
where $R_{\text{format}}(o_i)$ denotes a format reward (+0.5), requiring the model to place its reasoning within \texttt{<thinking></thinking>} tags and the final answer within \texttt{\textbackslash boxed\{\}}. This encourages the model to reason before answering and facilitates answer extraction. $R_{\text{accuracy}}(o_i)$ denotes an accuracy-based reward determined by the output correctness, providing positive feedback (+1) if the answer is correct and zero otherwise. 
Based on the rewards within the same rollout group, we compute the normalized advantage for each trajectory as follows:
\begin{equation}
A_i = \frac{r_i - \mathrm{mean}\{r_1, r_2, \dots, r_G\}}{\mathrm{std}\{r_1, r_2, \dots, r_G\}}.
\label{eq:advantage}
\end{equation}
We use maximizing-differential advantage-aware pruning to retain only the trajectories with the largest differences in gradient contributions. By prioritizing informative learning signals, the model is encouraged to focus on the most representative trajectories while reducing unnecessary gradient computation. 
We first partition trajectories according to their advantage values as follows:
\begin{equation}
S_k(q)=\{o_i \mid A_i=a_k\},
\label{eq:grouping}
\end{equation}
where $a_k$ denotes the $k$-th unique advantage value, and $S_k(q)$ represents the set of trajectories sharing the same advantage. This grouping aligns trajectories with similar dominant gradient tendencies into the same subspace, laying a foundation for redundancy reduction under consistent gradient semantic constraints. 
Empirical analysis of the correlation between token-level Jaccard similarity and the cosine similarity of actual gradients demonstrates a strong positive correlation among trajectories that share the same advantage within each question, as shown in Appendix \ref{sec:Gradient_Cosine_Similarity}. 
Additionally, token-level Jaccard similarity is computationally efficient and well-suited for pruning, we adopt it as a proxy within the same advantage group for gradient cosine similarity to characterize trajectory similarity, thereby serving as a prior estimate of gradient similarity. 
Specifically, for any two trajectories $o_i, o_j$ within the same group, we define their token-level Jaccard similarity for prior gradient contribution deviation as follows: 
\begin{equation}
\theta_{ij}=1-\frac{|N_n(o_i)\cap N_n(o_j)|}{|N_n(o_i)\cup N_n(o_j)|}.
\end{equation}
where $N_n(o_i)$ denotes the set of unique $n$-gram token sequences extracted from trajectory $o_i$. 
By estimating gradient similarity, highly similar trajectories exhibit smaller deviations in gradient contribution, whereas those with low similarity exhibit greater deviations. Thus, we prune trajectory pairs with $\theta_{ij}<\tau$ to remove unnecessary trajectories, including low-information trajectories and overly similar to already selected ones. Following this criterion, we iteratively construct a de-redundant subset within each advantage group. 
Specifically, starting from an empty set $S'_k(q)=\emptyset$, we traverse trajectories in $S_k(q)$ and prune any trajectory $o_i$ if there exists a previously retained trajectory $o_j \in S'_k(q)$ such that $\theta_{ij}<\tau$. Formally, the retained subset is defined as follows:
\begin{equation}
S'_k(q)=\{o_i \in S_k(q)\mid \not\exists\ o_j \in S'_k(q),\ \theta_{ij}<\tau\}.
\end{equation}
This process corresponds to a greedy maximal independent set selection under the similarity threshold, ensuring minimal overlap in gradient contributions between any two retained trajectories, while maximizing structural diversity within each group.
The final set of retained trajectories is given by:
\begin{equation}
S(q) = \mathop{\textstyle\bigcup}\limits_{k} S'_k(q).
\end{equation}
After maximizing differential advantage pruning, the remaining trajectories are fed into the old policy model, current policy model, and reference model. The hidden states are projected through linear layers to obtain logits, followed by a softmax over the vocabulary to compute token probabilities. These probabilities are then used to compute inter-model ratios for gradient updates.
By eliminating unnecessary trajectories, our proposed method significantly reduces the computational cost of forward and backward propagation, while alleviating policy bias and improving training stability.

\subsection{Adaptive Rollout Sampling}
The aforementioned pruning effectively reduces redundant computation at the gradient update stage, yet GRPO-family methods still suffer from an inherent efficiency limitation stemming from the rollout sampling scale. 
Compared with a static strategy that allocates a fixed yet potentially redundant number of trajectories for each question, we design an adaptive rollout sampling mechanism that dynamically adjusts the sampling scale based on historical pruning statistics. Thus, the information density of sampled trajectories can be readily inferred from pruning records. Specifically, heavy pruning corresponds to high redundancy and calls for a smaller rollout size, while high trajectory retention implies undersampling, requiring more trajectories for stable learning signals. 
Formally, given a question $q$ at training epoch $t$, we denote the number of sampled trajectories as $G_t^{\text{rollout}}(q)$ and the number used for gradient updates as $G_t^{\text{update}}(q)$, whose values remain identical across most GRPO variants. However, after applying Maximizing Differential Advantage Pruning, the actual number of trajectories used for updates becomes $G_t'(q)$, which satisfies $G_t^{\text{rollout}}(q) \geq G_t'(q)$, thereby reducing redundant computation.
To capture the overall pruning dynamics, we compute statistics of the retained update counts across all questions at epoch $t$:
\begin{equation}
G_t'^{\text{mean}} = \frac{1}{N} \sum_{i=1}^{N} G_t'(q_i), \quad
G_t'^{\text{max}} = \max_{i=1,\dots,N} G_t'(q_i),
\end{equation}
where $N$ is the number of questions in the current epoch. $G_t'^{\text{mean}}$ reflects the average information density, while $G_t'^{\text{max}}$ captures the upper bound of update counts required by the single question.
Based on these statistics, we define the composite pruning intensity as follows: 
\begin{equation}
G_t' = \alpha G_t'^{\text{mean}} + (1 - \alpha) G_t'^{\text{max}}, \quad
\alpha = \frac{\text{step}_{\text{cur}}}{\text{step}_{\text{total}}}.
\end{equation}
The $G_t'$ is defined over the range $[1, G_1]$.The coefficient $\alpha$ increases over training, encouraging exploration in the early stage (favoring $G_t'^{\text{max}}$) and gradually shifting toward efficiency (favoring $G_t'^{\text{mean}}$) as training progresses.
We further define a reference threshold as follows:
\begin{equation}
G_{\text{ref}}' = \frac{G_1'^{\text{mean}} + G_1'^{\text{max}}}{2},
\end{equation}
where $G_1'^{\text{mean}}$  denotes the average update count across all questions at the initial stage, and $G_1'^{\text{max}}$ denotes the maximum update count in a single question at the initial stage. Together, they provide a robust and generalizable baseline for dynamically adjusting the sampling scale throughout the training process.
Based on this threshold, we only trigger adjustment when the current pruning level falls below the reference (indicating high redundancy, requiring fewer rollouts) as follows:
\begin{equation}
\Delta_t =
\begin{cases}
G_{\text{ref}}' - G_t', & \text{if } G_t' < G_{\text{ref}}' \\
0, & \text{otherwise}
\end{cases}
\end{equation}
Finally, we adaptively adjust the rollout count for the next epoch as follows:
\begin{equation}
G_{t+1}^{\text{rollout}} =
\max\left(
\mathrm{round}(G_1-\Delta_t),\ 2
\right)
\end{equation}
where $G_1$ is the initial rollout count. Overall, the adaptive rollout sampling mechanism enables the model to dynamically adjust the trajectory sampling scale based on the observed pruning behavior, thereby improving computational efficiency while maintaining sufficient training signal quality throughout different training stages.

\section{Experiments}

\subsection{Experimental Settings}
\label{sec:experimental_settings}

\textbf{Datasets and Baseline Methods.} To evaluate the effectiveness of FastRL, we conduct training on the \textit{in-domain} datasets, Geometry3K~\cite{lu2021inter} and GeoQA8K-R1V\footnote{\href{https://huggingface.co/datasets/leonardPKU/GEOQA_8K_R1V}{https://huggingface.co/datasets/leonardPKU/GEOQA\_8K\_R1V}}, and further evaluate it on four \textit{out-of-domain} multimodal mathematical reasoning benchmarks, including MathVision~\cite{wang2024measuring}, MathVista~\cite{lu2023mathvista}, MathVerse~\cite{zhang2024mathverse}, and WeMath~\cite{qiao2025we}. These benchmarks cover diverse question types (e.g., geometry, chart, and table-based questions) with multi-level knowledge granularity, enabling comprehensive and fine-grained evaluation of visual mathematical reasoning. 
Dataset statistics are summarized in Table~\ref{tab:datasets}. 
\textbf{For baseline methods,} we compare FastRL with several representative methods from the GRPO family, including GRPO~\cite{shao2024deepseekmath}, DAPO~\cite{yu2025dapo}, and GSPO~\cite{zheng2025group}, as well as recent approaches designed to accelerate GRPO, such as CPPO~\cite{lin2025cppo} and GRESO~\cite{zheng2025act}. Specifically, CPPO applies global threshold-based pruning on advantage values across trajectories throughout training, reducing the computational cost of both forward and backward passes. GRESO maintains historical training information and incorporates curriculum learning to selectively filter out low-information simple questions and overly difficult questions, thereby reducing the number of questions involved in training. 

\begin{table*}[t]
\caption{Performance comparison (\%) on Geometry3K and GeoQA8K-R1V regarding accuracy. The \textbf{bold} numbers denote the best results. \textbf{Avg} denotes the average accuracy across out-of-domain datasets, and \textbf{Speed} denotes the training speed regarding GRPO.  $\uparrow$ indicates that the higher is better, while  $\downarrow$  indicates that the lower is better.}
\centering
\setlength{\tabcolsep}{4.0pt}
\scriptsize
\renewcommand{\arraystretch}{0.89}
\begin{tabular}{c|l|c|cccc|ccr}
\toprule
\textbf{Dataset} & \textbf{Method} & \textbf{In-domain} & \textbf{MathVerse} & \textbf{MathVision} & \textbf{MathVista} & \textbf{WeMath} & \textbf{Avg($\uparrow$)} & \textbf{Train-Time($\downarrow$)} & \textbf{Speed($\uparrow$)} \\
\midrule
\multicolumn{10}{c}{\cellcolor{lightgray} \textbf{Qwen2.5-VL-7B-Instruct}} \\
\midrule

\multirow{12}{*}{\rotatebox[origin=c]{90}{Geometry3K}}
& GRPO~\cite{shao2024deepseekmath}
& 53.74
& 43.38
& 26.25
& 66.90
& 68.74
& 51.32
& 29.72 & 1$\times$\\
& \hspace{0.5mm}+CPPO~\cite{lin2025cppo}
& 54.41
& 43.98
& 27.37
& 66.30
& 69.08
& 51.68
& 18.52 & 1.60$\times$\\
& \hspace{0.5mm}+GRESO~\cite{zheng2025act}
& 53.24
& 42.08
& 26.09
& 64.90
& 67.93
& 50.25
& 27.29 & 1.09$\times$ \\
& \cellcolor{lightblue}\hspace{0.4mm}+FastRL (Ours)
& \cellcolor{lightblue}\textbf{55.90}
& \cellcolor{lightblue}\textbf{45.76}
& \cellcolor{lightblue}\textbf{27.96}
& \cellcolor{lightblue}\textbf{67.30}
& \cellcolor{lightblue}\textbf{69.48}
& \cellcolor{lightblue}\textbf{52.63}
& \cellcolor{lightblue}\textbf{{13.87}} & \cellcolor{lightblue}\textbf{2.14$\times$} \\
\cmidrule(lr){2-10}

& DAPO~\cite{yu2025dapo}
& 53.57
& 41.24
& 26.25
& 67.20
& \textbf{70.69}
& 51.35
& {28.08} & 1$\times$ \\
& \hspace{0.5mm}+CPPO
& 54.57
& 43.58
& 27.30
& 68.40
& 69.94
& 52.31
& {16.61} & {1.69$\times$} \\
& \hspace{0.5mm}+GRESO
& 53.07
& 43.40
& 26.68
& 65.10
& 68.63
& 50.95
& {26.67} & {1.05$\times$} \\
& \cellcolor{lightblue}\hspace{0.5mm}+FastRL (Ours)
& \cellcolor{lightblue}\textbf{54.90}
& \cellcolor{lightblue}\textbf{43.93}
& \cellcolor{lightblue}\textbf{27.96}
& \cellcolor{lightblue}\textbf{69.80}
& \cellcolor{lightblue}70.63
& \cellcolor{lightblue}\textbf{53.08}
& \cellcolor{lightblue}\textbf{{13.29}} & \cellcolor{lightblue}\textbf{{2.11$\times$}} \\
\cmidrule(lr){2-10}

& GSPO~\cite{zheng2025group}
& 52.41
& 40.15
& 26.18
& 65.40
& 66.26
& 49.50
& 26.33 & 1$\times$ \\
& \hspace{0.5mm}+CPPO
& 53.24
& 41.27
& 26.09
& \textbf{66.60}
& 69.25
& 50.80
& 15.56 & {1.69$\times$}\\
& \hspace{0.5mm}+GRESO
& 52.58
& 39.92
& 25.00
& 65.30
& 68.10
& 49.58
& 25.33 & {1.04$\times$}\\
& \cellcolor{lightblue}\hspace{0.5mm}+FastRL (Ours)
& \cellcolor{lightblue}\textbf{54.57}
& \cellcolor{lightblue}\textbf{45.48}
& \cellcolor{lightblue}\textbf{26.97}
& \cellcolor{lightblue}66.40
& \cellcolor{lightblue}\textbf{70.57}
& \cellcolor{lightblue}\textbf{52.36}
& \cellcolor{lightblue}\textbf{{12.84}} & \cellcolor{lightblue}\textbf{{2.05$\times$}} \\

\midrule

\multirow{12}{*}{\rotatebox[origin=c]{90}{\centering GeoQA8K-R1V}}
& GRPO
& 68.30
& 45.18
& 26.78
& 68.60
& 68.68
& 52.31
& {54.65} & 1$\times$  \\
& \hspace{0.5mm}+CPPO
& 68.03
& 45.20
& 26.97
& 69.00
& 68.79
& 52.49
& {39.22} & {1.39$\times$}\\
& \hspace{0.5mm}+GRESO
& 68.70
& \textbf{45.99}
& 26.38
& 67.20
& 68.56
& 52.03
& {48.33} & {1.13$\times$}\\
& \cellcolor{lightblue}\hspace{0.5mm}+FastRL (Ours)
& \cellcolor{lightblue}\textbf{70.15}
& \cellcolor{lightblue}45.91
& \cellcolor{lightblue}\textbf{27.86}
& \cellcolor{lightblue}\textbf{70.60}
& \cellcolor{lightblue}\textbf{70.00}
& \cellcolor{lightblue}\textbf{53.59}
& \cellcolor{lightblue}\textbf{{25.65}} & \cellcolor{lightblue}\textbf{{2.13$\times$}}  \\
\cmidrule(lr){2-10}

& DAPO
& 67.90
& 43.96
& 25.99
& 68.30
& 66.67
& 51.23
& {53.22} & 1$\times$ \\
& \hspace{0.5mm}+CPPO
& 68.70
& 44.92
& 27.43
& 67.40
& 67.76
& 51.88
& {38.15}  & {1.39$\times$}\\
& \hspace{0.5mm}+GRESO
& 67.90
& 45.05
& 26.12
& 64.40
& 68.10
& 50.92
& {46.85} & {1.14$\times$}\\
& \cellcolor{lightblue}\hspace{0.5mm}+FastRL (Ours)
& \cellcolor{lightblue}\textbf{70.29}
& \cellcolor{lightblue}\textbf{46.17}
& \cellcolor{lightblue}\textbf{27.53}
& \cellcolor{lightblue}\textbf{68.50}
& \cellcolor{lightblue}\textbf{68.97}
& \cellcolor{lightblue}\textbf{52.79}
& \cellcolor{lightblue}\textbf{{24.62}} & \cellcolor{lightblue}\textbf{{2.16$\times$}} \\
\cmidrule(lr){2-10}

& GSPO
& 66.97
& 44.21
& 27.01
& 65.60
& \textbf{68.74}
& 51.39
& {52.87} & 1$\times$ \\
& \hspace{0.5mm}+CPPO
& 68.03
& 45.41
& 27.63
& \textbf{67.30}
& 68.56
& 52.23
& {37.55} & {1.41$\times$}\\
& \hspace{0.5mm}+GRESO
& 67.37
& 42.99
& 27.07
& 65.50
& 66.38
& 50.49
& {47.08} & {1.12$\times$}\\
& \cellcolor{lightblue}\hspace{0.5mm}+FastRL (Ours)
& \cellcolor{lightblue}\textbf{69.36}
& \cellcolor{lightblue}\textbf{46.27}
& \cellcolor{lightblue}\textbf{27.89}
& \cellcolor{lightblue}\textbf{67.30}
& \cellcolor{lightblue}68.45
& \cellcolor{lightblue}\textbf{52.48}
& \cellcolor{lightblue}\textbf{{23.85}}  & \cellcolor{lightblue}\textbf{{2.22$\times$}}\\
\midrule

\multicolumn{10}{c}{\cellcolor{lightgray} \textbf{Qwen3-VL-8B-Instruct}} \\
\midrule

\multirow{6}{*}{\rotatebox[origin=c]{90}{Geometry3K}}
& GRPO~\cite{shao2024deepseekmath}
& 78.87
& 55.76
& 48.98
& 71.20
& 80.63
& 64.14
& 30.05 & {1$\times$} \\

& DAPO~\cite{yu2025dapo} 
& 79.53
& 55.61
& 50.39
& \textbf{71.60}
& 80.06
& 64.42
& 28.67 & {1.05$\times$} \\

& GSPO~\cite{zheng2025group} 
& 79.03
& 55.28
& 49.67
& 71.10
& 80.69
& 64.19
& 28.32 & {1.06$\times$}  \\

& CPPO~\cite{lin2025cppo}
& 79.36
& 55.03
& 51.45
& 70.70
& 78.79
& 63.99
& 22.75 & {1.32$\times$} \\
& GRESO ~\cite{zheng2025act}
& 78.87
& 54.82
& 50.53
& 70.20
& 78.22
& 63.44
& 29.11& {1.03$\times$}  \\
& \cellcolor{lightblue}FastRL (Ours)
& \cellcolor{lightblue}\textbf{80.03}
& \cellcolor{lightblue}\textbf{56.45}
& \cellcolor{lightblue}\textbf{54.14}
& \cellcolor{lightblue}71.10
& \cellcolor{lightblue}\textbf{81.61}
& \cellcolor{lightblue}\textbf{65.83}
& \cellcolor{lightblue}\textbf{16.37} & \cellcolor{lightblue}\textbf{{1.84$\times$}} \\

\midrule

\multirow{6}{*}{\rotatebox[origin=c]{90}{\centering GeoQA8k-R1V}}
& GRPO
& 82.36
& 54.01
& 49.05
& 70.40
& 78.22
& 62.92
& {85.16} & {1$\times$}\\

& DAPO
& 82.36
& 53.58
& 50.59
& 70.00
& 79.31
& 63.37
& {85.01} & {1.00$\times$}\\

& GSPO
& 82.49
& 54.31
& 49.57
& 71.50
& \textbf{79.77}
& 63.79
& {83.25} & {1.02$\times$}\\

& CPPO
& 84.08
& 52.64
& 46.64
& 69.80
& 76.00
& 61.27
& {68.22} & {1.25$\times$}\\
& GRESO
& 82.09
& 52.74
& 47.86
& 69.70
& 77.01
& 61.83
& {84.67} & {1.01$\times$} \\
& \cellcolor{lightblue}FastRL (Ours)
& \cellcolor{lightblue}\textbf{84.35}
& \cellcolor{lightblue}\textbf{55.36}
& \cellcolor{lightblue}\textbf{52.20}
& \cellcolor{lightblue}\textbf{71.60}
& \cellcolor{lightblue}78.85
& \cellcolor{lightblue}\textbf{64.50}
& \cellcolor{lightblue}\textbf{{45.33}}  & \cellcolor{lightblue}\textbf{{1.88$\times$}} \\

\bottomrule
\end{tabular}
\label{tab:main_results}
\vspace{-0.45cm}
\end{table*}

\textbf{Implementation Details.} 
We implement FastRL and baselines using the EasyR1~\cite{zheng2025easyr1} framework, and extend GRESO and CPPO to the multimodal setting for comparison. 
Considering the scale of Geometry3K and GeoQA8K-R1V, the number of training epochs is set to 25 and 15, respectively. We adopt the AdamW optimizer with a learning rate of $1\times10^{-6}$ and a weight decay of $1\times10^{-2}$. The KL divergence coefficient $\beta$ is set to 0.01.
For all methods, the initial rollout sampling number $G_1$ is fixed to 8, the training batch size is set to 256, the hyperparameter $\tau$ is set to 0.8, and the remaining hyperparameters follow the default settings of EasyR1. All experiments are conducted on NVIDIA H20 GPUs, and  Qwen2.5-VL-7B-Instruct and Qwen3-VL-8B-Instruct are trained on four and eight GPUs, respectively. We use \textbf{accuracy (\%)} and \textbf{training time (h)} as evaluation metrics. 

\subsection{Overall Results and Analysis }
As shown in Table~\ref{tab:main_results}, FastRL consistently outperforms all baselines regarding both training efficiency and average accuracy across Geometry3K and GeoQA8K-R1V using both Qwen2.5-VL-7B-Instruct and Qwen3-VL-8B-Instruct. Notably, FastRL (with Qwen2.5-VL-7B-Instruct) accelerates GRPO training by up to 2.14$\times$ and 2.22$\times$ on the Geometry3K and GeoQA8K-R1V benchmarks, respectively. Meanwhile, it improves average accuracy by 1.31\% and 1.09\% on these two benchmarks, respectively. Similar trends are observed with Qwen3-VL-8B-Instruct, further validating the robustness of FastRL across model scales. These improvements stem primarily from maximizing differential advantage pruning and adaptive rollout sampling, which significantly reduce redundant computation while preserving discriminative trajectories, thereby balancing efficiency and performance. 

Compared with GRPO, CPPO achieves training acceleration but yields limited, or even worse, accuracy across different backbones and datasets. For example, CPPO (with Qwen3-VL-8B-Instruct) accelerates GRPO training while reducing average accuracy by 0.15\% and 1.65\% on Geometry3K and GeoQA8K-R1V, respectively. This suggests that pruning trajectories by a fixed ratio may remove informative trajectories that remain valuable for policy optimization, thereby affecting training stability. Furthermore, CPPO's speedup decreases from 1.60× to 1.39× when transferring to GeoQA8K-R1V, likely because CPPO only reduces optimization-stage computation while leaving rollout unchanged. Since longer responses increase the relative cost of rollout, the overall speedup becomes limited. Note that GRESO provides limited acceleration at the cost of performance, as its curriculum-based filtering introduces overhead for maintaining historical statistics while risking premature removal of informative questions, ultimately harming generalization. In contrast, FastRL achieves consistent speedup across both backbones and datasets while maintaining or improving accuracy, demonstrating a better balance between efficiency and performance. 

\subsection{Ablation Study}

\begin{table*}[t]
\caption{Ablation study (\%) of FastRL. “w/o MDAP” denotes FastRL without maximizing differential advantage pruning (with ARS simulated to match its standard dynamics for a fair comparison), and “w/o ARS” denotes FastRL without adaptive rollout sampling.}
\centering
\setlength{\tabcolsep}{2.5pt}
\scriptsize
\renewcommand{\arraystretch}{0.8}
\begin{tabular}{l|l|l|cccc|lcl}
\toprule
\textbf{Dataset} & \textbf{Method} & \textbf{In-domain} 
& \textbf{MathVerse} & \textbf{MathVision} & \textbf{MathVista} & \textbf{WeMath} & \textbf{Avg($\uparrow$)} & \textbf{Train-Time($\downarrow$)} & \textbf{Speed($\uparrow$)} \\
\midrule

\multicolumn{10}{c}{\cellcolor{lightgray} \textbf{Qwen2.5-VL-7B-Instruct}} \\
\midrule

\multirow{3}{*}{Geometry3K}
& FastRL
& 55.90
& 45.76 & 27.96 & 67.30 & 69.48 & 52.63 &13.87&2.14$\times$  \\
& w/o MDAP
& 53.41\reddrop{2.49}
&45.00	&27.04	&66.30	&69.31	&51.91\reddrop{0.72} &25.85	 &1.15$\times$\reddrop{0.99$\times$} \\
& w/o ARS
&54.01\reddrop{1.89}		&45.00	&27.99	&67.10	&68.97	&52.27\reddrop{0.36} &14.41&2.06$\times$\reddrop{0.08$\times$} \\

\cmidrule(lr){1-10}

\multirow{3}{*}{GeoQA8K-R1V}
& FastRL
& 70.15
& 45.91 & 27.86 & 70.60 & 70.00 & 53.59 &25.65 &2.13$\times$ \\
& w/o MDAP
&68.70\reddrop{1.45}	&44.95	&27.43	&67.60	&69.54	&52.38\reddrop{1.21} &49.75 &1.10$\times$\reddrop{1.03$\times$} \\
& w/o ARS
&69.62\reddrop{0.53}	&45.38	&27.80	&70.00	&69.83	&53.25\reddrop{0.34} &29.53 &1.85$\times$\reddrop{0.28$\times$}\\

\midrule

\multicolumn{10}{c}{\cellcolor{lightgray} \textbf{Qwen3-VL-8B-Instruct}} \\
\midrule

\multirow{3}{*}{Geometry3K}
& FastRL
& 80.03
& 56.45 & 54.14 & 71.10 & 81.61 & 65.83 &16.37  &1.84$\times$\\
& w/o MDAP
&78.70\reddrop{1.33} 
&55.56  &51.74  &70.20  &80.11  &64.40\reddrop{1.43}  &27.42 &1.10$\times$\reddrop{0.74$\times$} \\
& w/o ARS
&79.70\reddrop{0.33} 
&56.07  &53.85  &70.60  &80.69  &65.30\reddrop{0.53}  &17.97 &1.67$\times$\reddrop{0.17$\times$} \\

\cmidrule(lr){1-10}

\multirow{3}{*}{GeoQA8K-R1V}
& FastRL
&84.35 
&55.36  &52.20  &71.60  &78.85  &64.50  &45.33 &1.88$\times$ \\
& w/o MDAP
&82.89\reddrop{1.46} 
&53.73  &51.32  &70.10  &78.22  &63.34\reddrop{1.16}  &79.25 &1.07$\times$\reddrop{0.81$\times$} \\
& w/o ARS
&83.55\reddrop{0.80} 
&54.92  &52.14  &71.20  &79.14  &64.35\reddrop{0.15}  &50.12  &1.70$\times$\reddrop{0.18$\times$}\\

\bottomrule
\end{tabular}
\label{tab:ablation}
\end{table*}

We conduct ablation studies on both Qwen2.5-VL-7B-Instruct and Qwen3-VL-8B-Instruct to assess the contributions of the two core components in FastRL. As shown in Table~\ref{tab:ablation}, removing the maximizing differential advantage pruning (MDAP) leads to the most significant degradation in both accuracy and training efficiency. For instance, on Qwen2.5-VL-7B-Instruct, the average accuracy drops by 0.72\% and 1.21\%, while the training speed decreases sharply from 2.14$\times$ to 1.15$\times$ and 2.13$\times$ to 1.10$\times$ on Geometry3K and GeoQA8K-R1V, respectively, indicating that MDAP is the primary driver of performance and acceleration gains. In contrast, removing adaptive rollout sampling (ARS) results in smaller but consistent declines in both metrics, suggesting its complementary role in further boosting efficiency and stabilizing training. Such consistent trends across model scales further validate that our components complement each other and collectively enhance final performance. 

\begin{figure*}[t]
\centering
\includegraphics[width=1\textwidth]{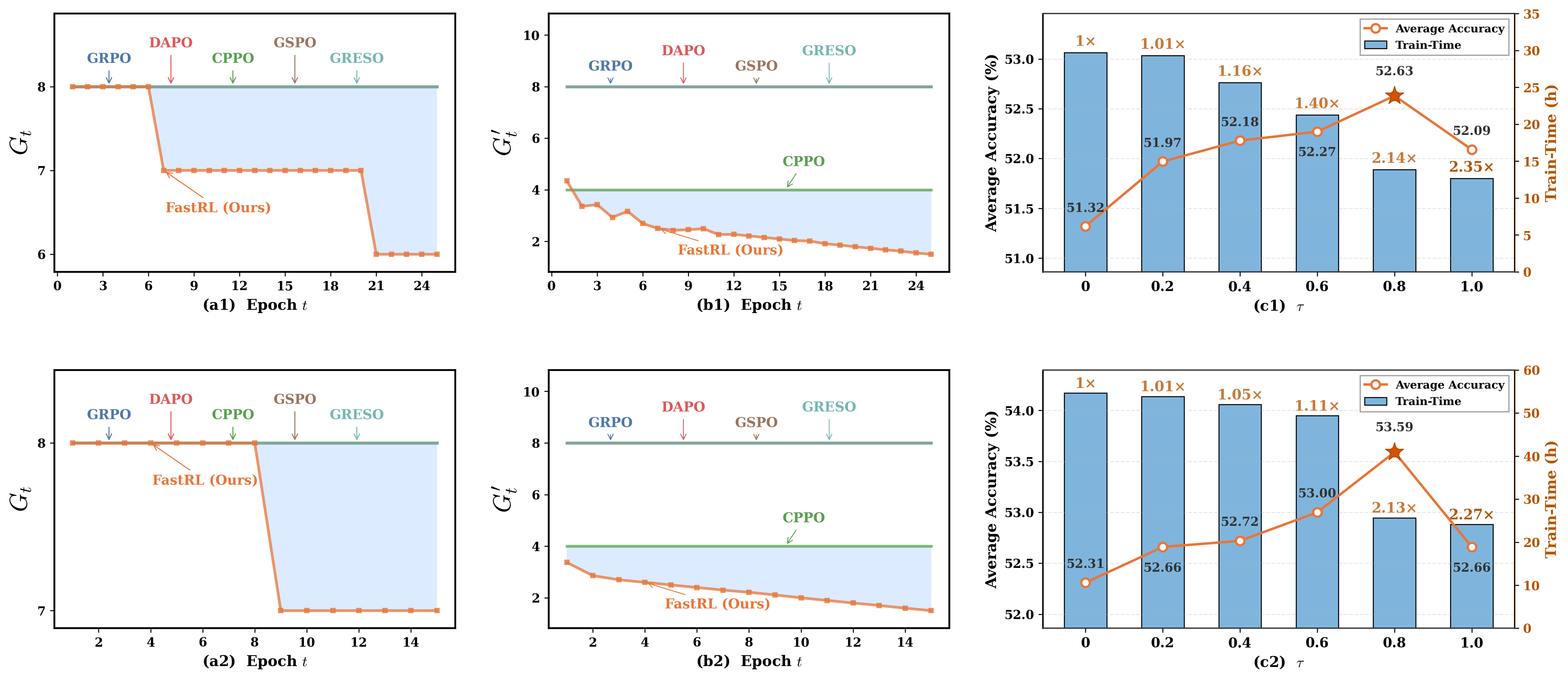}
\caption{
Training efficiency and parameter sensitivity of FastRL on Geometry3K (top) and GeoQA8K-R1V (bottom). 
(a1)-(a2) show the the rollout sampling scale $G_t$;
(b1)-(b2) present the number of trajectories used for policy updates; 
(c1)-(c2) show the sensitivity to the hyperparameter $\tau$.}
\label{fig:param_sensitivity}
\vspace{-0.5cm}
\end{figure*}
\begin{table*}[t]
\caption{Generalization across model architectures and datasets.}
\centering
\setlength{\tabcolsep}{3.5pt}
\scriptsize
\renewcommand{\arraystretch}{0.8}
\begin{tabular}{c|l|c|cccc|ccc}
\toprule
\textbf{Dataset} & \textbf{Method} & \textbf{In-domain} & \textbf{AIME2023} & \textbf{AIME2024} & \textbf{AIME2025} & \textbf{AIME2026} & \textbf{Avg($\uparrow$)} & \textbf{Train-Time($\downarrow$)} & \textbf{Speed($\uparrow$)} \\
\midrule
\multicolumn{10}{c}{\cellcolor{lightgray} \textbf{Llama3.1-8B-Instruct}} \\
\midrule

\multirow{6}{*}{\rotatebox[origin=c]{90}{MATH}}
& GRPO~\cite{shao2024deepseekmath}
&72.32 &\textbf{10.00} &10.00 &\textbf{6.67} &\textbf{3.33} &\textbf{7.50} & {21.16} & 1$\times$ \\

& DAPO~\cite{yu2025dapo} 
&72.72 &6.67 &10.00 &\textbf{6.67} &\textbf{3.33} & 6.67
& {19.72} & {1.07$\times$ } \\

& GSPO~\cite{zheng2025group} 
&72.56 &6.67 &10.00 &0.00 &0.00 &4.17 & {18.89} & {1.12$\times$ }  \\

& CPPO~\cite{lin2025cppo}
&71.92 &3.33 &10.00 &0.00 &\textbf{3.33} &4.17 & {11.44} & {1.85$\times$ } \\
& GRESO ~\cite{zheng2025act}
&71.60 &3.33 &6.67 &3.33 &0.00 &3.33 & {17.35}& {1.22$\times$ }  \\
& \cellcolor{lightblue}FastRL (Ours)
& \cellcolor{lightblue}\textbf{73.04}
& \cellcolor{lightblue}\textbf{10.00}
& \cellcolor{lightblue}\textbf{13.33}
& \cellcolor{lightblue}3.33
& \cellcolor{lightblue}\textbf{3.33}
& \cellcolor{lightblue}\textbf{7.50}
& \cellcolor{lightblue}\textbf{{9.22}} & \cellcolor{lightblue}\textbf{{2.30$\times$ }} \\

\bottomrule
\end{tabular}
\label{tab:Other_modalities}
\end{table*}

\subsection{Analysis of FastRL's Training Efficiency and Parameter Sensitivity}

\textbf{Training Efficiency Analysis.} As shown in Figure \ref{fig:param_sensitivity} (a1)-(a2), FastRL adaptively adjusts the rollout sampling scale based on the historical pruning distribution. The results indicate that this adaptive sampling mechanism achieves a favorable balance between exploration adequacy and computational efficiency. 
In addition, as illustrated in Figure \ref{fig:param_sensitivity} (b1)-(b2), $G_t'$ shows a gradual downward trend as training proceeds, indicating that the number of highly distinct trajectories gradually diminishes in the later training stages. 
Compared with CPPO, which adopts a similar advantage-based pruning strategy, FastRL retains fewer yet more representative trajectories for gradient updates. This leads to both improved training efficiency and enhanced model performance. 

\textbf{Parameter Sensitivity Analysis.} As shown in Figure~\ref{fig:param_sensitivity} (c1)-(c2), we perform trajectory pruning among rajectories with identical advantages by preferentially removing those with smaller deviations, that is, higher similarity. When the threshold $\tau$ is set to 0.8, FastRL achieves the highest benchmark accuracy, while attaining training speedups of 2.14$\times$ and 2.13$\times$ on these two datasets, respectively. When $\tau$ is further increased to 1.0, meaning that only one trajectory is retained for each advantage, the training speedups improve to 2.35$\times$ and 2.27$\times$, respectively, but the benchmark accuracy decreases compared to the case of $\tau = 0.8$. Thus, we set  $\tau=0.8$ as our final configuration since average accuracy peaks at this setting, achieving a good trade-off between performance and training efficiency. 

\subsection{Generalization Across Architectures and Datasets}

To evaluate the generalization of FastRL beyond its primary setting, we conduct experiments on Llama3.1-8B-Instruct, trained on the text-only MATH dataset~\cite{hendrycks2021measuring}, and compare against all baselines under identical conditions. As shown in Table~\ref{tab:Other_modalities}, FastRL achieves the highest in-domain accuracy (73.04\%) and the greatest training speedup ($2.30\times$), outperforming CPPO. On out-of-domain AIME benchmarks, it matches GRPO's average accuracy (7.50\%) with less than half the training time (9.22 vs.\ 21.16 hours), while efficiency-oriented methods such as CPPO and GRESO suffer notable accuracy degradation (Avg: 4.17\% and 3.33\%, respectively). These results confirm that FastRL's speedup does not compromise generalization, and its consistent gains across both a distinct backbone and a text-only dataset attest to its broad applicability. 

\begin{table*}[t]
\caption{Pass@16 results (\%) on out-of-domain benchmarks. Base model: Qwen2.5-VL-7B-Instruct.}
\centering
\setlength{\tabcolsep}{10.5pt}
\scriptsize
\renewcommand{\arraystretch}{0.8}
\begin{tabular}{c|l|cccc|c}
\toprule
\textbf{Dataset} & \textbf{Method} & \textbf{MathVerse} & \textbf{MathVision} & \textbf{MathVista} & \textbf{WeMath} & \textbf{avg@16($\uparrow$)} \\
\midrule

\multirow{6}{*}{Geometry3K}
& GRPO~\cite{shao2024deepseekmath}
& 60.53 & 62.96 & 82.50 & 90.34 & 74.08 \\
& DAPO~\cite{yu2025dapo}
& 58.98 & 62.57 & 81.50 & 90.57 & 73.41 \\
& GSPO~\cite{zheng2025group}
& 59.21 & 62.30 & 81.90 & 90.11 & 73.38 \\
& GRESO~\cite{zheng2025act}
& 59.44 & 62.47 & 82.00 & 90.23 & 73.54 \\
& CPPO~\cite{lin2025cppo}
& 60.81 & 63.19 & 82.20 & 90.69 & 74.22 \\
& \cellcolor{lightblue}FastRL (Ours)
& \cellcolor{lightblue}\textbf{63.60}
& \cellcolor{lightblue}\textbf{67.86}
& \cellcolor{lightblue}\textbf{83.60}
& \cellcolor{lightblue}\textbf{93.85}
& \cellcolor{lightblue}\textbf{77.23} \\

\cmidrule(lr){1-7}

\multirow{6}{*}{GeoQA8K-R1V}
& GRPO~\cite{shao2024deepseekmath}
& 60.99 & 59.41 & 82.10 & 89.77 & 73.07 \\
& DAPO~\cite{yu2025dapo}
& 59.57 & 57.86 & 81.90 & 87.59 & 71.73 \\
& GSPO~\cite{zheng2025group}
& 60.53 & 58.22 & 82.00 & 87.64 & 72.10 \\
& GRESO~\cite{zheng2025act}
& 59.26 & 57.24 & 81.50 & 86.78 & 71.20 \\
& CPPO~\cite{lin2025cppo}
& 60.41 & 58.88 & 82.00 & 89.08 & 72.59 \\
& \cellcolor{lightblue}FastRL (Ours)
& \cellcolor{lightblue}\textbf{62.39}
& \cellcolor{lightblue}\textbf{63.95}
& \cellcolor{lightblue}\textbf{83.20}
& \cellcolor{lightblue}\textbf{91.32}
& \cellcolor{lightblue}\textbf{75.21} \\

\bottomrule
\end{tabular}
\label{tab:pass16}
\end{table*}
\begin{table*}[t]
\caption{Ablation (\%) of pruning strategies on Geometry3K using Qwen2.5-VL-7B-Instruct.}
\centering
\setlength{\tabcolsep}{11pt}
\scriptsize
\renewcommand{\arraystretch}{0.8}
\begin{tabular}{l|cccc|c|c}
\toprule
\textbf{Pruning Strategy} & \textbf{MathVerse} & \textbf{MathVision} & \textbf{MathVista} & \textbf{WeMath} & \textbf{Avg($\uparrow$)} & \textbf{Speed($\uparrow$)} \\
\midrule
Random-pruning 0.25 & 43.98 & 27.40 & 66.90 & 68.85 & 51.78 & 1.27$\times$ \\
Random-pruning 0.50 & 45.35 & 26.74 & 66.60 & 67.70 & 51.59 & 1.60$\times$ \\
Random-pruning 0.75 & 42.79 & 26.45 & 64.60 & 66.49 & 50.08 & 2.10$\times$ \\
Advantage-only (Ours) & 44.54 & 27.57 & 67.10 & 69.14 & 52.09 & \textbf{2.35$\times$} \\
\rowcolor{lightblue}
FastRL (Ours) & \textbf{45.76} & \textbf{27.96} & \textbf{67.30} & \textbf{69.48} & \textbf{52.63} & 2.14$\times$ \\
\bottomrule
\end{tabular}
\label{tab:pruning_ablation}
\end{table*}

\subsection{Further Analysis}

\textbf{Exploration Capacity Analysis.} To verify the impact of trajectory pruning on the exploration space of the policy, we adopt Pass@$k$, a widely used metric for estimating the upper bound of a model's capability. As shown in Table~\ref{tab:pass16}, FastRL does not shrink the exploration space or cap the potential capability of the model. On the contrary, it improves avg@16 over GRPO by 3.15 on Geometry3K and 2.14 on GeoQA8K-R1V, and attains the best Pass@16 score on every individual benchmark. This indicates that pruning redundant trajectories that share identical advantages does not suppress exploration, since such trajectories provide little additional learning signal for policy optimization.

\textbf{Analysis of Pruning Strategies.} We further conduct a targeted ablation to disentangle the two sources of gain in FastRL: (i) \textit{diversity preservation across advantage groups}, and (ii) \textit{redundancy pruning within the same advantage group based on Jaccard distance}. Specifically, we compare the following settings on Geometry3K using Qwen2.5-VL-7B-Instruct: \textit{Random-pruning} 0.25/0.5/0.75, which randomly prunes 25\%/50\%/75\% of the trajectories within each query; \textit{Advantage-only}, which groups trajectories by advantage and randomly retains one trajectory per group without Jaccard pruning; and \textit{FastRL}, the full method that additionally applies Jaccard-distance-based pruning to redundant trajectories within each advantage group.
As shown in Table~\ref{tab:pruning_ablation}, we draw the following conclusions. (1) Random pruning yields acceleration but suffers from significant performance degradation as the pruning ratio increases, showing that blindly reducing the number of trajectories is not the source of the gain. (2) Simply grouping by advantage and retaining one trajectory per group (\textit{Advantage-only}) already achieves a $2.35\times$ speedup together with a slight performance improvement, indicating that most of the acceleration and the stable gain comes from removing redundant updates with identical advantages. (3) Adding Jaccard-based pruning on top of advantage grouping (the full FastRL) further improves the average accuracy from 52.09 to 52.63, showing that Jaccard-distance pruning plays the key role of preventing high-value trajectories that share the same reward but differ substantially in structure from being mistakenly pruned within the same advantage group, and is therefore a critical component for balancing speed and performance. In summary, the speedup of FastRL mainly comes from redundancy compression across advantage groups, while Jaccard-distance pruning ensures that structurally diverse trajectories carrying additional learning signals are preserved during compression. Together, they form the favorable efficiency-performance trade-off of FastRL.

\section{Conclusion}

This paper proposes FastRL, a plug-and-play and efficient reinforcement learning framework that improves training efficiency while enhancing policy learning. Specifically, we introduce a structured pruning strategy based on maximizing advantage diversity, which retains only the most informative trajectories with the largest differences in gradient contribution, allowing the model to focus on highly discriminative learning signals while reducing redundant forward computation. Furthermore, we develop an adaptive rollout sampling mechanism coupled with pruning dynamics, which adjusts the sampling scale online according to the historical pruning distribution, achieving a balance between exploration sufficiency and computational efficiency. Experimental results demonstrate that FastRL consistently delivers both training acceleration and accuracy improvements for GRPO-based methods. 

\begin{ack}
This work is partially supported by the research project funded by Guangdong Basic and Applied Basic Research Foundation (2026A1515011829, 2024A1515140144), the Fundamental Research Funds for the Central Universities (21624325, 21624338, 21625102), Ministry of Education of the People's Republic of China Humanities and Social Sciences Youth Foundation (24YJC890034), the Key Laboratory of Smart Education of Guangdong Higher Education Institute, Jinan University (2022LSYS003). 
This work is also supported by NSFC (62377028), Guangdong Science and Technology Plan Project (2025A0505010018), "Master Mentor Plan" of Jinan University (YDXS2501), and the teaching reform research projects of Jinan University (JG2026030). We also thank the OPPO AI Center for providing computational resources for this work.

\end{ack}

\bibliographystyle{plainnat}
\bibliography{nips}

\newpage
\appendix
\begin{center}
    \LARGE \textbf{Appendix}
\end{center}

\section{Experimental Datasets. }
\begin{figure*}[h]
\centering
\includegraphics[width=1\textwidth]{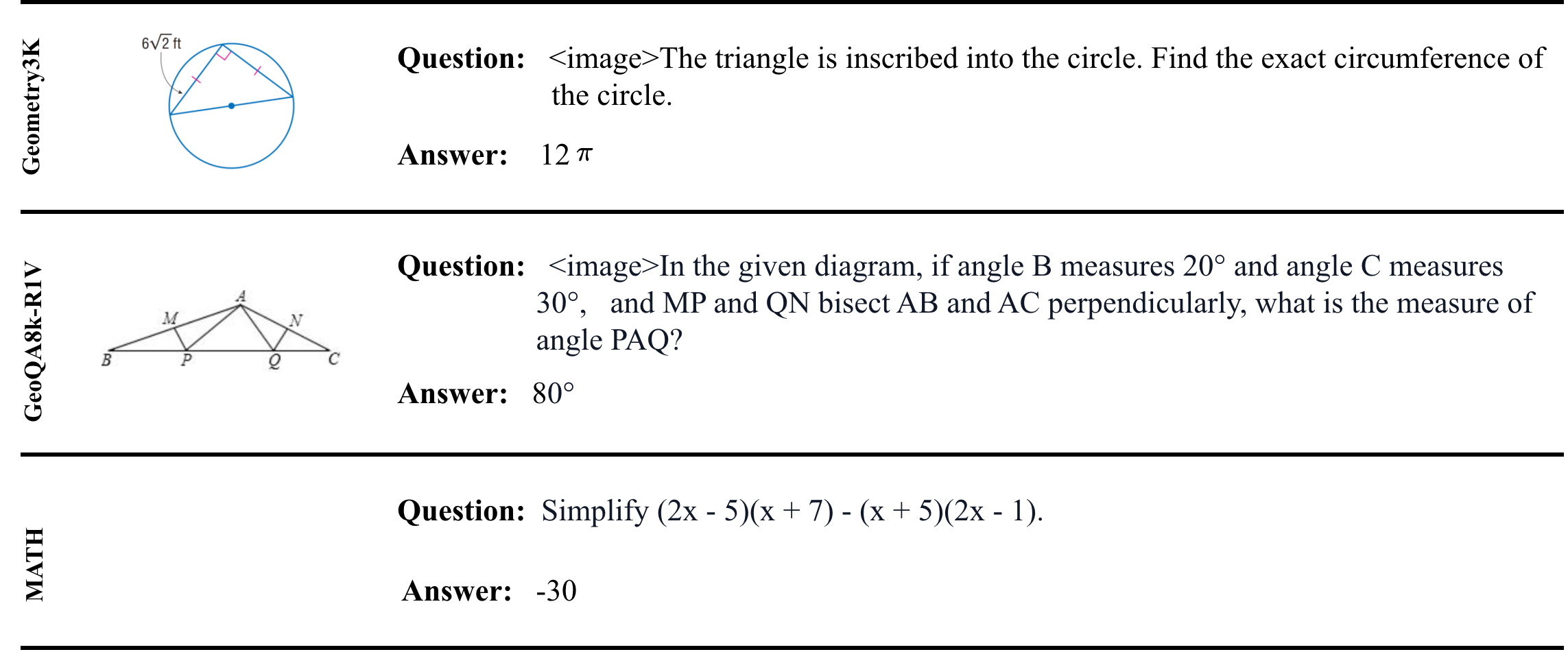}
\caption{Examples from the three training datasets: Geometry3K, GeoQA8k-R1V, and MATH.}
\label{fig:dataset_example}
\end{figure*}

To validate the effectiveness of our method, we conduct experiments in both multimodal and unimodal settings. We first provide a brief overview of in-domain datasets. 
The Geometry3K~\cite{lu2021inter} dataset contains 3,002 questions, and we use 2,702 problems from its training and test splits in our experiments, covering geometry knowledge from grades 6 to 12 in North America. GeoQA8K-R1V\footnote{\href{https://huggingface.co/datasets/leonardPKU/GEOQA_8K_R1V}{https://huggingface.co/datasets/leonardPKU/GEOQA\_8K\_R1V}} is a multimodal geometric reasoning dataset for reinforcement learning, specifically designed for the R1-V paradigm to enhance the geometric reasoning capabilities of vision-language models (VLMs).
The MATH~\cite{hendrycks2021measuring} dataset contains 12,500 highly challenging text-based mathematical problems. Each problem in MATH is accompanied by a complete step-by-step solution, which can be used to train models to generate reasoning processes and detailed explanations alongside final answers. Examples are illustrated in Figure~\ref{fig:dataset_example}. 
In addition, we provide a brief overview of out-of-domain benchmarks used to evaluate the models' reasoning ability. 
MathVision~\cite{wang2024measuring} is a challenging benchmark consisting of 3,040 mathematical questions with visual context, collected from real-world math competitions across 12 grade levels from primary school to high school.
MathVista~\cite{lu2023mathvista} is a comprehensive benchmark for evaluating mathematical reasoning in visual contexts. It contains 1,000 questions spanning diverse question types, including geometry, charts, and tables.
MathVerse~\cite{zhang2024mathverse} is a comprehensive visual mathematics benchmark designed for a fair and in-depth evaluation of multimodal large language models. Its test set includes 3,940 multi-subject math questions with diagrams from publicly available sources.
WeMath~\cite{qiao2025we} carefully collects and organizes 1,740 visual math questions in its test set, covering 67 hierarchical knowledge concepts across five levels of granularity.
AIME\footnote{\href{https://maa.org/maa-invitational-competitions/}{https://maa.org/maa-invitational-competitions/}} is a text-only benchmark for evaluating advanced mathematical reasoning. It is designed to assess the multi-step reasoning capabilities of large language models. 

\begin{table}[h]
\caption{Statistics of the datasets used in our experiments.}
\setlength{\tabcolsep}{3.8pt}
\renewcommand{\arraystretch}{0.89}
\centering
\small
\begin{tabular}{lcccc|lcccc}
\toprule
\multicolumn{5}{c|}{\textbf{Multimodal}} & \multicolumn{5}{c}{\textbf{Unimodal}} \\
\cmidrule(r){1-5} \cmidrule(l){6-10}
\textbf{Dataset} & \textbf{Sum} & \textbf{Train} & \textbf{Test} & \textbf{Modality} 
& \textbf{Dataset} & \textbf{Sum} & \textbf{Train} & \textbf{Test} & \textbf{Modality} \\
\midrule
Geometry3K     & 2,702 & 2,101  & 601  & Text + Image 
& MATH       & 12,500 & 11,250 & 1,250 & Text \\
GeoQA8K-R1V    & 8,785 & 8,031  & 754  & Text + Image  
& AIME2023   & 30     & -      & 30    & Text \\
MathVerse      & 3,940 & -      & 3,940 & Text + Image  
& AIME2024   & 30     & -      & 30    & Text \\
MathVision     & 3,040 & -      & 3,040 & Text + Image  
& AIME2025   & 30     & -      & 30    & Text \\
MathVista (mini)& 1,000 & -      & 1,000 & Text + Image  
& AIME2026   & 30     & -      & 30    & Text \\
WeMath         & 1,740 & -      & 1,740 & Text + Image  
&             &       &        &       & \\
\bottomrule
\end{tabular}
\label{tab:datasets}
\end{table}

\section{Additional Method Details and Analysis}
\label{sec:Additional Method Details}

\subsection{Algorithm}

\begin{algorithm}[H]
\caption{FastRL: Maximizing Differential Advantage Pruning with Adaptive Rollout Sampling for GRPO}
\label{alg:fastrl}
\begin{algorithmic}[1]

\Require Initial model $\pi_{\theta_{\text{init}}}$; dataset $\mathcal{D}$; batch size $b$; group size $G$; similarity threshold $\tau$; hyperparameters $\epsilon$, $\beta$; total epochs $E$; total steps $M$
\State Policy model $\pi_\theta \leftarrow \pi_{\theta_{\text{init}}}$; \quad Reference model $\pi_{\text{ref}} \leftarrow \pi_\theta$
\State $n \leftarrow G$; \quad $G_{\text{ref}}' \leftarrow \texttt{None}$; \quad $G_t' \leftarrow \texttt{None}$ \hfill $\triangleright$ \textit{Adaptive sampling state}
\State $\mathcal{B}_{G_t'^{\text{mean}}} \leftarrow []$; \quad $\mathcal{B}_{G_t'^{\text{max}}} \leftarrow []$ \hfill $\triangleright$ \textit{Epoch-level buffers}

\For{step $= 1, \ldots, M$}

    \Statex \textcolor{red}{\textbf{\# Phase 1(a): Adaptive Rollout Sampling}}
    \If{$G_t' \neq \texttt{None}$ \textbf{and} $G_{\text{ref}}' \neq \texttt{None}$ \textbf{and} $G_t' < G_{\text{ref}}'$}
        \State $n \leftarrow \max\!\big(\text{round}(\min(G - (G_{\text{ref}}' - G_t'),\; G)),\; 2\big)$
    \Else
        \State $n \leftarrow G$
    \EndIf

    \State $\pi_{\theta_{\text{old}}} \leftarrow \pi_\theta$
    \State Sample batch $\mathcal{D}_b$ of $b$ questions from $\mathcal{D}$
    \State Sample $n$ trajectories $O(q) = \{o_1, \ldots, o_n\} \sim \pi_{\theta_{\text{old}}}(\cdot \mid q)$ for each $q \in \mathcal{D}_b$
    \State Compute rewards $\{r_i\}$ and advantages $\{A_i\}$ via Eq.~\ref{eq:advantage}

\Statex \textcolor{red}{\textbf{\# Phase 2: Maximizing Differential Advantage Pruning}}
    \For{each question $q \in \mathcal{D}_b$}
        \State Partition trajectories into groups $S_1(q), \ldots, S_K(q)$ by unique advantage values $a_k$
        \State $S(q) \leftarrow \emptyset$
        \For{each group $S_k(q)$}
            \State $S'_k(q) \leftarrow \emptyset$
            \For{each $o_i \in S_k(q)$ (ordered by $|A_i|$ descending)}
                \If{$\nexists\; o_j \in S'_k(q)$ s.t. $\theta_{ij} = 1 - \frac{|N_n(o_i)\cap N_n(o_j)|}{|N_n(o_i)\cup N_n(o_j)|} < \tau$}
                    \State $S'_k(q) \leftarrow S'_k(q) \cup \{o_i\}$ \hfill $\triangleright$ \textit{Keep diverse trajectory}
                \EndIf
            \EndFor
            \State $S(q) \leftarrow S(q) \cup S'_k(q)$ \hfill $\triangleright$ \textit{$|S'_k(q)| \geq 1$ guaranteed}
        \EndFor
    \EndFor

    \State $\mathcal{S} \leftarrow \mathop{\textstyle\bigcup}_{q} S(q)$
    \While{$|\mathcal{S}| \bmod d \neq 0$} \hfill $\triangleright$ \textit{$d$: divisor for DP alignment}
        \State Add unselected trajectory with highest $|A_i|$ via round-robin
    \EndWhile

    \Statex \textcolor{red}{\textbf{\# Phase 1(b): Adaptive Rollout Sampling Update}}
    \State $G_t'^{\text{mean}} \leftarrow \text{mean}(\{|S(q)|\})$; \quad $G_t'^{\text{max}} \leftarrow \max(\{|S(q)|\})$
    \State Append $G_t'^{\text{mean}}$ to $\mathcal{B}_{G_t'^{\text{mean}}}$; \quad Append $G_t'^{\text{max}}$ to $\mathcal{B}_{G_t'^{\text{max}}}$
    \If{new epoch boundary is reached}
        \State $\alpha \leftarrow \text{step} / M$
        \State 
        $G_t' \leftarrow \alpha \cdot \overline{\mathcal{B}}_{G_t'^{\text{mean}}} + (1 - \alpha) \cdot \overline{\mathcal{B}}_{G_t'^{\text{max}}}$
        \If{first epoch completed \textbf{and} $G_{\text{ref}}' = \texttt{None}$}
            \State 
            $G_{\text{ref}}' \leftarrow \left( \overline{\mathcal{B}}_{G_t'^{\text{mean}}} + \overline{\mathcal{B}}_{G_t'^{\text{max}}} \right) / 2$
        \EndIf
        \State Reset $\mathcal{B}_{G_t'^{\text{mean}}}, \mathcal{B}_{G_t'^{\text{max}}} \leftarrow [], []$
    \EndIf

    \State Update $\pi_\theta$ on selected trajectories $\mathcal{S}$

\EndFor

\Ensure $\pi_\theta$

\end{algorithmic}
\end{algorithm}

\subsection{Correlation Analysis of Token-Level Jaccard Similarity and Gradient Cosine Similarity}
\label{sec:Gradient_Cosine_Similarity}

\begin{figure*}[h]
\centering
\includegraphics[width=1\textwidth]{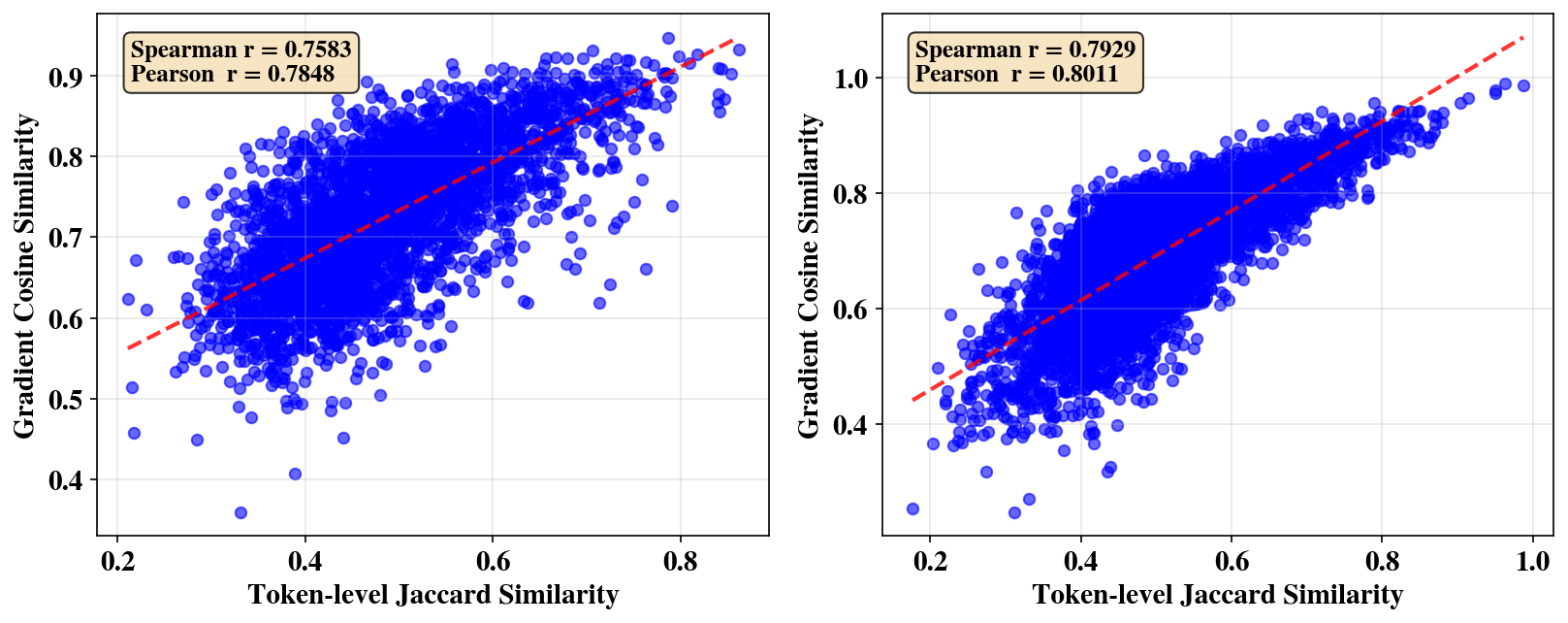}
\caption{Correlation analysis between token-level Jaccard similarity and gradient cosine similarity of trajectories with the same advantage for each question. The left plot corresponds to the Geometry3K dataset, while the right plot corresponds to the GeoQA8k-R1V dataset. The high Pearson and Spearman correlation coefficients demonstrate a strong positive correlation between the two metrics across both datasets.}
\label{fig:jaccard_vs_gradient_correlation}
\end{figure*}

To validate the correlation between token level Jaccard similarity and gradient cosine similarity among trajectories that share the same advantage for each question, we conduct an empirical study during reinforcement learning training with the GRPO algorithm on the Geometry3K and GeoQA8K-R1V datasets using Qwen2.5-VL-7B-Instruct. Specifically, for each dataset, we randomly sample 256 questions across different training epochs, yielding a total of 2048 trajectories. For each question, we select trajectory pairs with the same advantage and compute their token-level Jaccard similarity and gradient cosine similarity, then perform a correlation analysis between the two metrics.
As shown in Figure \ref{fig:jaccard_vs_gradient_correlation}, both datasets exhibit a strong positive correlation between these two measures, with high Pearson and Spearman coefficients. This observation indicates that trajectories with more similar token-level structures tend to produce more aligned policy gradient directions. These results provide empirical evidence that token-level Jaccard similarity serves as an effective proxy for gradient similarity, thereby supporting its use in the pruning strategy proposed in this work.

\section{Additional Experimental Results}

\subsection{Training Efficiency under Low Policy Lag.}
\label{app:low_policy_lag}
\begin{figure*}[t]
\centering
\includegraphics[width=1\textwidth]{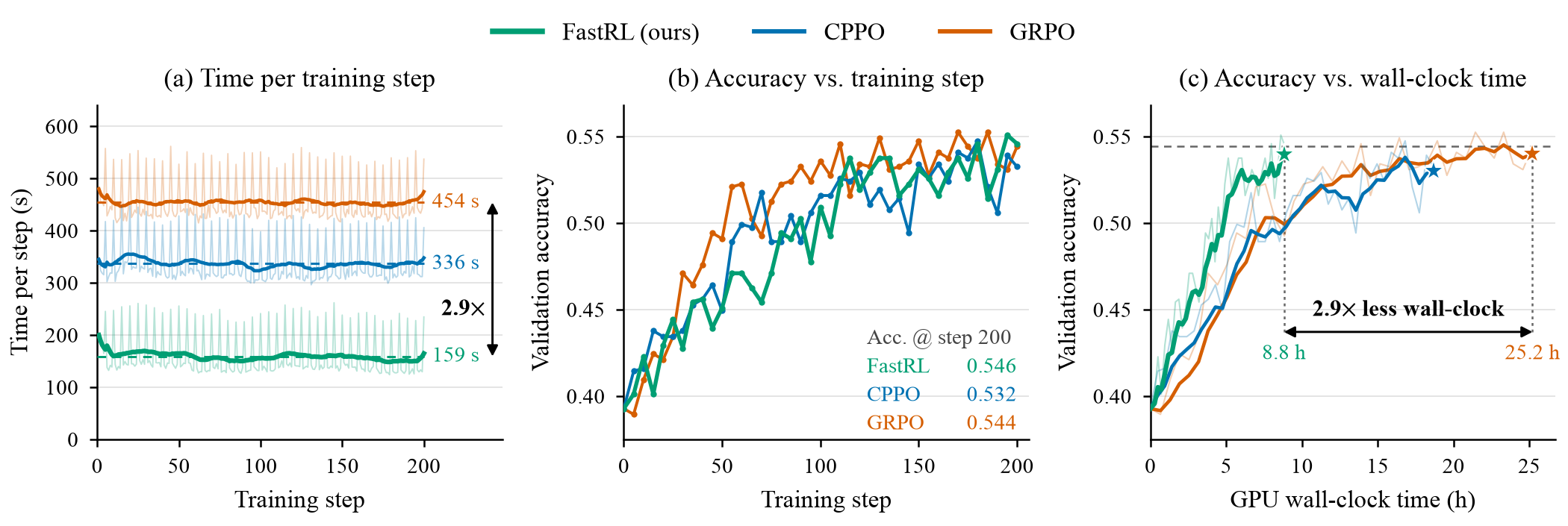}
\caption{Training efficiency of Qwen2.5-VL-7B-Instruct on Geometry3K under the low policy lag setting (training batch size 256, mini-batch size 64, i.e., 4 policy updates per training step). (a) Time per training step (thin: raw; thick: moving average; dashed: mean). (b) Validation accuracy \textit{vs.} training step. (c) Validation accuracy \textit{vs.} cumulative GPU wall-clock time.}
\label{fig:low_policy_lag}
\end{figure*}

In GRPO-style training, the rollouts of each training step are split into several mini-batches on which the policy is updated sequentially, so the later updates are performed on data generated by an increasingly outdated policy. We refer to the number of policy updates per training step, i.e., the ratio between the training batch size and the mini-batch size, as the \textit{policy lag}. We first consider the commonly adopted low policy lag setting, where the training batch size is 256 and the mini-batch size is 64, yielding 4 policy updates per step. Since CPPO and FastRL reduce the number of trajectories used for optimization, we also fix their number of updates per step to 4 for a fair comparison (see the update-count alignment below). As shown in Figure~\ref{fig:low_policy_lag}(a), FastRL takes only 159\,s per step on average, compared with 336\,s for CPPO and 454\,s for GRPO, i.e., $2.9\times$ cheaper per step than GRPO. Meanwhile, Figure~\ref{fig:low_policy_lag}(b) shows that the peak accuracy achieved by FastRL within 200 training steps is comparable to that of GRPO and higher than that of CPPO. Consequently, as shown in Figure~\ref{fig:low_policy_lag}(c), FastRL completes 200 training steps in only 8.8\,h, whereas CPPO and GRPO require 18.7\,h and 25.2\,h, respectively, achieving a $2.9\times$ wall-clock speedup over GRPO without sacrificing accuracy. Since the number of policy updates is aligned across methods, these gains stem from pruning redundant trajectories and adaptively reducing the rollout scale, rather than from performing fewer optimization steps.

\textbf{Further optimizations of FastRL.} Note that the speedup achieved here ($2.9\times$) is higher than that reported on Geometry3K in Table~\ref{tab:main_results} ($2.14\times$). This is because the experiments in this subsection adopt an improved version of FastRL, in which we introduce two further optimizations after the main experiments. Both optimizations leave the pruning criterion and the optimization objective unchanged, and only remove the remaining redundant computation.

(i) \textit{Aligning the number of updates instead of the mini-batch size.} In the version used in the main experiments, the mini-batch size is fixed, so the number of retained trajectories after pruning must be divisible by the mini-batch size (i.e., the number of prompts per mini-batch multiplied by $G_1$). To satisfy this constraint, part of the pruned trajectories have to be backfilled according to their absolute advantages, which re-introduces redundant trajectories into the update stage; moreover, the number of policy updates per step then varies with the pruning ratio and becomes inconsistent with that of the baselines. In the improved version, we instead fix the number of policy updates per step and evenly split the retained trajectories into the corresponding number of mini-batches, whose size is dynamically determined by the actual number of retained trajectories. Consequently, the retained batch only needs to be divisible by the number of GPUs for data-parallel dispatching, so that the backfilling overhead becomes negligible, and the number of optimization steps is kept identical to that of the baselines.

(ii) \textit{Pruning before computing log-probabilities.} The version used in the main experiments follows the default pipeline of EasyR1, i.e., old-policy log-probabilities $\rightarrow$ reference log-probabilities $\rightarrow$ advantage estimation $\rightarrow$ pruning $\rightarrow$ policy update, where the forward passes of the old policy and the reference model are performed on the full rollout batch. However, the group-relative advantage in Eq.~\ref{eq:advantage} depends only on rewards, and our pruning criterion depends only on advantages and token-level Jaccard similarity, neither of which requires log-probabilities. We therefore reorder the pipeline as advantage estimation $\rightarrow$ pruning $\rightarrow$ old-policy log-probabilities $\rightarrow$ reference log-probabilities $\rightarrow$ policy update, so that both forward passes are performed only on the retained trajectories. Since the KL divergence is incorporated as a loss term rather than a reward penalty and no critic is involved, this reordering leaves the optimization objective unchanged and only removes the redundant forward computation on pruned trajectories.
Together, these two optimizations remove redundant computation in both the log-probability computation and the policy update stages, thereby further enlarging the end-to-end speedup of FastRL.

\subsection{Training Dynamics and Computational Overhead under High Policy Lag.}
\label{app:high_policy_lag}
\begin{figure*}[t]
\centering
\includegraphics[width=1\textwidth]{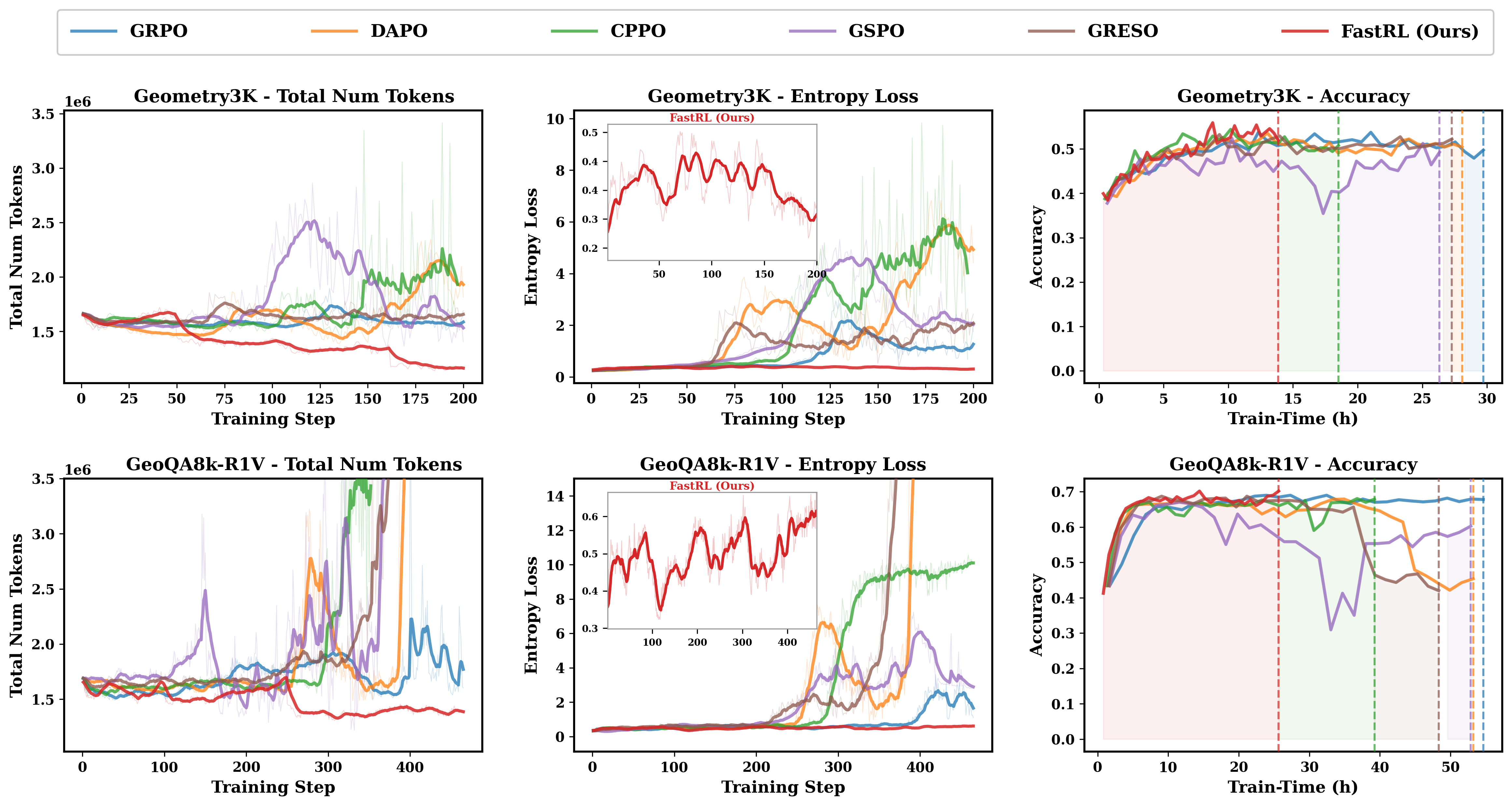}
\caption{Training dynamics and computational overhead analysis of Qwen2.5-VL-7B-Instruct on Geometry3K and GeoQA8K-R1V under the high policy lag setting (training batch size 256, mini-batch size 16, i.e., 16 policy updates per training step).
We compare FastRL (ours) with GRPO, DAPO, CPPO, GSPO, and GRESO across three key metrics: total token consumption (left), policy entropy (middle), and accuracy (right).}
\label{fig:total_token_entroy_loss_acc}
\end{figure*}

To further examine the robustness of FastRL when optimization becomes strongly off-policy, we increase the policy lag by reducing the mini-batch size to 16 while keeping the training batch size at 256. In this setting, each training step performs 16 sequential policy updates, and the later updates rely on increasingly stale rollouts, which places a much heavier burden on training stability. Figure~\ref{fig:total_token_entroy_loss_acc} presents a comprehensive comparison between FastRL and several representative GRPO-style baselines on the Geometry3K and GeoQA8K-R1V datasets using Qwen2.5-VL-7B-Instruct under this setting. Specifically, we analyze three key aspects throughout training, including total token consumption, policy entropy evolution, and accuracy with respect to wall-clock training time.

First, as shown in the left column of Figure~\ref{fig:total_token_entroy_loss_acc}, FastRL consistently maintains the lowest total token consumption across the entire training process on both datasets. In contrast, under high policy lag, existing baselines generally exhibit progressively increasing token usage, accompanied by substantial fluctuations in later training stages, suggesting that the accumulated off-policy drift gradually pushes the policy toward verbose and unstable generations. This observation demonstrates that FastRL effectively reduces computational overhead while maintaining stable training behavior. The improvement mainly stems from the synergy between the maximum-difference advantage pruning strategy and the adaptive rollout sampling mechanism. Specifically, the pruning strategy removes trajectories with highly redundant gradient contributions, while the adaptive sampling mechanism dynamically adjusts the rollout scale according to historical pruning statistics, thereby preserving sufficient exploration capability while avoiding unnecessary sampling costs.

Second, the middle column illustrates the evolution of policy entropy during training. Under high policy lag, repeated updates on stale rollouts make the current policy deviate increasingly from the behavior policy that generated the data, and existing GRPO-style methods become prone to severe entropy instability: their entropy remains low in the early stage but surges abruptly during the middle and later stages, followed by highly unstable oscillatory or degenerative behaviors. Such behavior indicates that the policy drifts beyond the region where the clipped surrogate objective remains reliable, which leads to unstable optimization dynamics and degraded generalization performance. In contrast, FastRL maintains a consistently stable entropy throughout training, with moderate fluctuations that reflect sustained exploration. This phenomenon suggests that the proposed method remains robust even under strongly off-policy optimization. We attribute this improvement to the proposed maximum-difference advantage pruning mechanism, which removes highly homogeneous trajectories that would otherwise be repeatedly reinforced across the many sequential updates, and explicitly preserves trajectories with diverse advantage patterns and gradient contributions, enabling the policy to continuously learn from heterogeneous optimization signals instead of overfitting to stale and redundant ones.

Finally, the right column presents the relationship between accuracy and wall-clock training time. FastRL achieves competitive or superior performance while requiring substantially less training time compared with all baselines. More importantly, FastRL reaches stable high accuracy significantly earlier than existing methods, demonstrating markedly improved optimization efficiency. In contrast, several baselines (e.g., DAPO, GSPO, and GRESO on GeoQA8K-R1V) suffer from pronounced accuracy degradation in the later stages of training, which coincides with the entropy surge in the middle column, indicating that high policy lag substantially undermines their training stability. Even for baselines that eventually approach comparable performance, considerably higher computational cost is required.

Overall, combined with the results under low policy lag in Appendix~\ref{app:low_policy_lag}, these results consistently demonstrate that FastRL achieves a favorable trade-off between computational efficiency, training stability, and final task performance across different degrees of policy lag. FastRL consistently accelerates training in both regimes, and under high policy lag it further mitigates the entropy instability and performance degradation that commonly affect existing GRPO-style methods, while preserving strong exploration capability throughout reinforcement learning training.

\subsection{Supplementary Parameter Sensitivity Analysis.}

\begin{table*}[h]
\caption{Supplementary Parameter Sensitivity Analysis.}
\centering
\setlength{\tabcolsep}{3pt}
\scriptsize
\begin{tabular}{l|l|c|cccc|ccr}
\toprule
\textbf{Dataset} & \textbf{$\tau$} & \textbf{In-domain} 
& \textbf{MathVerse} & \textbf{MathVision} & \textbf{MathVista} & \textbf{WeMath} & \textbf{Avg($\uparrow$)} & \textbf{Train-Time($\downarrow$)} & \textbf{Speed($\uparrow$)} \\
\midrule

\multicolumn{10}{c}{\cellcolor{lightgray} \textbf{Qwen2.5-VL-7B-Instruct}} \\
\midrule

\multirow{6}{*}{Geometry3K}
& 0(GRPO)
& 53.74
& 43.38
& 26.25
& 66.90
& 68.74
& 51.32 &29.72  &1$\times$  \\
& 0.2
&54.08	&44.37	&27.37	&67.10 &69.02&51.97&29.33&1.01$\times$  \\
& 0.4
&54.58  &44.80	&\textbf{27.99}	&67.00	&68.68	&52.18	&25.64 &1.16$\times$ \\
& 0.6
&55.24 &44.75	&28.13	&66.90	&69.31	&52.27 &21.26 &1.40$\times$ \\
& 0.8
&\textbf{55.90}	&\textbf{45.76}	&27.96	&\textbf{67.30}	&\textbf{69.48}	&\textbf{52.63} &13.87 &2.14$\times$\\
& 1.0
&54.41		&44.54	&27.57	&67.10	&69.14	&52.09 &\textbf{12.65}  &\textbf{2.35$\times$} \\
\cmidrule(lr){1-10}

\multirow{6}{*}{GeoQA8K-R1V}
& 0(GRPO)
& 68.30
& 45.18
& 26.78
& 68.60
& 68.68
& 52.31  &54.65  &1$\times$  \\
& 0.2
&68.43  &44.67	&27.80	&69.10	&69.08	&52.66 &53.87&1.01$\times$ \\
& 0.4
&68.70	&45.23	&27.63	&69.50 &68.51	& 52.72&51.98&1.05$\times$ \\
& 0.6
&69.23 &45.05	&\textbf{27.86}	&70.40	&68.68	&53.00 &49.39 &1.11$\times$\\
& 0.8
&\textbf{70.15}	&\textbf{45.91}	&\textbf{27.86}	&\textbf{70.60}	&\textbf{70.00}	&\textbf{53.59} &25.65& 2.13$\times$ \\
& 1.0
&68.91		&44.39	&27.04	&69.90	&69.31	&52.66 &\textbf{24.11}&\textbf{2.27$\times$}\\

\midrule
\midrule

\textbf{Dataset} & \textbf{$\tau$} & \textbf{In-domain} 
& \textbf{AIME2023} & \textbf{AIME2024} & \textbf{AIME2025} & \textbf{AIME2026} & \textbf{Avg($\uparrow$)} & \textbf{Train-Time($\downarrow$)} & \textbf{Speed($\uparrow$)} \\
\midrule

\multicolumn{10}{c}{\cellcolor{lightgray} \textbf{Llama3.1-8B-Instruct}} \\
\midrule

\multirow{6}{*}{MATH}
& 0(GRPO)
&72.32 &10.00 &10.00 &6.67&3.33 &7.50 & {21.16} & 1$\times$ \\
& 0.2
&72.48  &6.67  &13.33  &3.33 &3.33  &6.67  &20.67  & 1.02$\times$  \\
& 0.4
&72.80  &6.67  &13.33  &3.33  &3.33  &6.67  &19.32  & 1.10$\times$ \\
& 0.6
&72.96  &10.00  &13.33  &3.33  &3.33  &7.50  &16.50  & 1.28$\times$ \\
& 0.8
&73.04  &10.00  &13.33  &3.33   &3.33   &7.50  &9.22  &2.30$\times$ \\
& 1.0
&72.72  &10.00  &10.00  &3.33  &3.33  &6.67  &9.02  &2.35$\times$ \\

\bottomrule
\end{tabular}
\label{tab:Parameter}
\end{table*}

We conduct a systematic sensitivity analysis of the hyperparameter $\tau$ on multimodal datasets Geometry3K and GeoQA8K-R1V using Qwen2.5-VL-7B-Instruct, as well as on the pure-text dataset MATH using Llama3.1-8B-Instruct. The detailed numerical results are reported in Table~\ref{tab:Parameter}. As $\tau$ increases, the training time is substantially reduced and training efficiency is significantly improved. Meanwhile, the average accuracy across in-domain and benchmark evaluations consistently improves, reaching its best performance at $\tau = 0.8$.
This trend can be attributed to our method’s ability to maximize differential advantages by retaining only trajectories with the most significant gradient contributions. This encourages the model to focus on the most discriminative learning signals, while maintaining sufficient exploration and reducing unnecessary forward and backward computations. In addition, the incorporation of adaptive rollout sampling further enhances both performance and training efficiency.

\subsection{Extension to Continuous-Reward Settings.}
\label{sec:continuous_reward}

\begin{table*}[h]
\caption{FastRL vs. GRPO on the grounding task under continuous rewards. The base model is Qwen3.5-9B.}
\centering
\setlength{\tabcolsep}{11pt}
\scriptsize
\renewcommand{\arraystretch}{1.2}
\begin{tabular}{l|cc|cc|cc}
\toprule
\textbf{Method} & \multicolumn{2}{c|}{\textbf{General(en)}} & \multicolumn{2}{c|}{\textbf{General(cn)}} & \multicolumn{2}{c}{\textbf{Efficiency}} \\
\cmidrule(lr){2-3} \cmidrule(lr){4-5} \cmidrule(lr){6-7}
& \textbf{center-in} & \textbf{mean-IoU} & \textbf{center-in} & \textbf{mean-IoU} & \textbf{Train-Time (h)($\downarrow$)} & \textbf{Speed($\uparrow$)} \\
\midrule
GRPO~\cite{shao2024deepseekmath} & 95.58 & 89.03 & 94.04 & 87.19 & 49.21 & 1.00$\times$ \\
\rowcolor{lightblue}
FastRL (Ours) & \textbf{97.33} & \textbf{89.19} & \textbf{94.91} & \textbf{88.45} & \textbf{28.01} & \textbf{1.76$\times$} \\
\bottomrule
\end{tabular}
\label{tab:grounding}
\end{table*}

 To further evaluate the broader applicability of FastRL, we apply it to a grounding task, where the model must output the bounding box of the target region given a query. We sample 50,000 examples from the open-source OS-Atlas dataset\footnote{\href{https://huggingface.co/datasets/OS-Copilot/OS-Atlas-data}{https://huggingface.co/datasets/OS-Copilot/OS-Atlas-data}} as the training set, and adopt three reward functions: two continuous rewards (an IoU reward and a center-in reward) and one discrete reward (a format reward). For the continuous rewards, we only adjust the reward/advantage grouping strategy in Eq.~\eqref{eq:grouping}: two trajectories $o_i$ and $o_j$ are assigned to the same advantage group whenever $|A_i - A_j| < \epsilon$ with $\epsilon = 10^{-4}$, so that the infinitesimal numerical differences introduced by continuous rewards do not fragment the advantage groups.
Under this setting, we compare GRPO and FastRL with Qwen3.5-9B on 32 NVIDIA H20 GPUs using the Verl training framework, with Megatron as the training engine. The batch size is 128 and the mini-batch size is 32, both the prompt and response lengths are capped at 10K tokens, and all other hyper-parameters follow the Verl defaults. As shown in Table~\ref{tab:grounding}, even under continuous rewards FastRL still achieves a $1.76\times$ training speedup over GRPO, while delivering more competitive performance on both the English and Chinese general benchmarks. This shows that FastRL is not restricted to discrete mathematical reasoning and can be extended to broader RL training scenarios.

\section{Limitations and Future Work}
\label{sec:limitation}
Maximizing Differential Advantage Pruning (MDAP) is primarily designed for reinforcement learning settings with discrete reward signals. Although this assumption covers most practical applications, extending MDAP to continuous reward functions still requires appropriate modifications to the reward representation. A possible approach is to discretize continuous rewards using piecewise functions as an approximation. However, this process typically depends on task-specific design choices and requires careful tuning.

\newpage
\section*{NeurIPS Paper Checklist}


\begin{enumerate}

\item {\bf Claims}
    \item[] Question: Do the main claims made in the abstract and introduction accurately reflect the paper's contributions and scope?
    \item[] Answer: \answerYes{} 
    \item[] Justification: Our abstract and introduction accurately reflect this paper’s contributions and scope.
    \item[] Guidelines:
    \begin{itemize}
        \item The answer \answerNA{} means that the abstract and introduction do not include the claims made in the paper.
        \item The abstract and/or introduction should clearly state the claims made, including the contributions made in the paper and important assumptions and limitations. A \answerNo{} or \answerNA{} answer to this question will not be perceived well by the reviewers. 
        \item The claims made should match theoretical and experimental results, and reflect how much the results can be expected to generalize to other settings. 
        \item It is fine to include aspirational goals as motivation as long as it is clear that these goals are not attained by the paper. 
    \end{itemize}

\item {\bf Limitations}
    \item[] Question: Does the paper discuss the limitations of the work performed by the authors?
    \item[] Answer: \answerYes{} 
    \item[] Justification:  Please refer to Sec.~\ref{sec:limitation} in the Appendix for details. 
    \item[] Guidelines:
    \begin{itemize}
        \item The answer \answerNA{} means that the paper has no limitation while the answer \answerNo{} means that the paper has limitations, but those are not discussed in the paper. 
        \item The authors are encouraged to create a separate ``Limitations'' section in their paper.
        \item The paper should point out any strong assumptions and how robust the results are to violations of these assumptions (e.g., independence assumptions, noiseless settings, model well-specification, asymptotic approximations only holding locally). The authors should reflect on how these assumptions might be violated in practice and what the implications would be.
        \item The authors should reflect on the scope of the claims made, e.g., if the approach was only tested on a few datasets or with a few runs. In general, empirical results often depend on implicit assumptions, which should be articulated.
        \item The authors should reflect on the factors that influence the performance of the approach. For example, a facial recognition algorithm may perform poorly when image resolution is low or images are taken in low lighting. Or a speech-to-text system might not be used reliably to provide closed captions for online lectures because it fails to handle technical jargon.
        \item The authors should discuss the computational efficiency of the proposed algorithms and how they scale with dataset size.
        \item If applicable, the authors should discuss possible limitations of their approach to address problems of privacy and fairness.
        \item While the authors might fear that complete honesty about limitations might be used by reviewers as grounds for rejection, a worse outcome might be that reviewers discover limitations that aren't acknowledged in the paper. The authors should use their best judgment and recognize that individual actions in favor of transparency play an important role in developing norms that preserve the integrity of the community. Reviewers will be specifically instructed to not penalize honesty concerning limitations.
    \end{itemize}

\item {\bf Theory assumptions and proofs}
    \item[] Question: For each theoretical result, does the paper provide the full set of assumptions and a complete (and correct) proof?
    \item[] Answer: \answerYes{} 
    \item[] Justification: Please refer to sec.~\ref{sec:Preliminaries} and Appendix sec.~\ref{sec:Additional Method Details}
    \item[] Guidelines:
    \begin{itemize}
        \item The answer \answerNA{} means that the paper does not include theoretical results. 
        \item All the theorems, formulas, and proofs in the paper should be numbered and cross-referenced.
        \item All assumptions should be clearly stated or referenced in the statement of any theorems.
        \item The proofs can either appear in the main paper or the supplemental material, but if they appear in the supplemental material, the authors are encouraged to provide a short proof sketch to provide intuition. 
        \item Inversely, any informal proof provided in the core of the paper should be complemented by formal proofs provided in appendix or supplemental material.
        \item Theorems and Lemmas that the proof relies upon should be properly referenced. 
    \end{itemize}

    \item {\bf Experimental result reproducibility}
    \item[] Question: Does the paper fully disclose all the information needed to reproduce the main experimental results of the paper to the extent that it affects the main claims and/or conclusions of the paper (regardless of whether the code and data are provided or not)?
    \item[] Answer: \answerYes{} 
    \item[] Justification: Please refer to Sec.~\ref{sec:experimental_settings}. Code is available via an anonymous link. 
    \item[] Guidelines:
    \begin{itemize}
        \item The answer \answerNA{} means that the paper does not include experiments.
        \item If the paper includes experiments, a \answerNo{} answer to this question will not be perceived well by the reviewers: Making the paper reproducible is important, regardless of whether the code and data are provided or not.
        \item If the contribution is a dataset and\slash or model, the authors should describe the steps taken to make their results reproducible or verifiable. 
        \item Depending on the contribution, reproducibility can be accomplished in various ways. For example, if the contribution is a novel architecture, describing the architecture fully might suffice, or if the contribution is a specific model and empirical evaluation, it may be necessary to either make it possible for others to replicate the model with the same dataset, or provide access to the model. In general. releasing code and data is often one good way to accomplish this, but reproducibility can also be provided via detailed instructions for how to replicate the results, access to a hosted model (e.g., in the case of a large language model), releasing of a model checkpoint, or other means that are appropriate to the research performed.
        \item While NeurIPS does not require releasing code, the conference does require all submissions to provide some reasonable avenue for reproducibility, which may depend on the nature of the contribution. For example
        \begin{enumerate}
            \item If the contribution is primarily a new algorithm, the paper should make it clear how to reproduce that algorithm.
            \item If the contribution is primarily a new model architecture, the paper should describe the architecture clearly and fully.
            \item If the contribution is a new model (e.g., a large language model), then there should either be a way to access this model for reproducing the results or a way to reproduce the model (e.g., with an open-source dataset or instructions for how to construct the dataset).
            \item We recognize that reproducibility may be tricky in some cases, in which case authors are welcome to describe the particular way they provide for reproducibility. In the case of closed-source models, it may be that access to the model is limited in some way (e.g., to registered users), but it should be possible for other researchers to have some path to reproducing or verifying the results.
        \end{enumerate}
    \end{itemize}

\item {\bf Open access to data and code}
    \item[] Question: Does the paper provide open access to the data and code, with sufficient instructions to faithfully reproduce the main experimental results, as described in supplemental material?
    \item[] Answer: \answerYes{} 
    \item[] Justification: We provide an anonymous link for reproducing our method. 
    \item[] Guidelines:
    \begin{itemize}
        \item The answer \answerNA{} means that paper does not include experiments requiring code.
        \item Please see the NeurIPS code and data submission guidelines (\url{https://neurips.cc/public/guides/CodeSubmissionPolicy}) for more details.
        \item While we encourage the release of code and data, we understand that this might not be possible, so \answerNo{} is an acceptable answer. Papers cannot be rejected simply for not including code, unless this is central to the contribution (e.g., for a new open-source benchmark).
        \item The instructions should contain the exact command and environment needed to run to reproduce the results. See the NeurIPS code and data submission guidelines (\url{https://neurips.cc/public/guides/CodeSubmissionPolicy}) for more details.
        \item The authors should provide instructions on data access and preparation, including how to access the raw data, preprocessed data, intermediate data, and generated data, etc.
        \item The authors should provide scripts to reproduce all experimental results for the new proposed method and baselines. If only a subset of experiments are reproducible, they should state which ones are omitted from the script and why.
        \item At submission time, to preserve anonymity, the authors should release anonymized versions (if applicable).
        \item Providing as much information as possible in supplemental material (appended to the paper) is recommended, but including URLs to data and code is permitted.
    \end{itemize}

\item {\bf Experimental setting/details}
    \item[] Question: Does the paper specify all the training and test details (e.g., data splits, hyperparameters, how they were chosen, type of optimizer) necessary to understand the results?
    \item[] Answer: \answerYes{} 
    \item[] Justification: Please refer to sec.~\ref{sec:experimental_settings}
    \item[] Guidelines:
    \begin{itemize}
        \item The answer \answerNA{} means that the paper does not include experiments.
        \item The experimental setting should be presented in the core of the paper to a level of detail that is necessary to appreciate the results and make sense of them.
        \item The full details can be provided either with the code, in appendix, or as supplemental material.
    \end{itemize}

\item {\bf Experiment statistical significance}
    \item[] Question: Does the paper report error bars suitably and correctly defined or other appropriate information about the statistical significance of the experiments?
    \item[] Answer: \answerNo{} 
    \item[] Justification: Due to the high training cost, the main experiments were conducted with a single run. Nevertheless, we demonstrate the robustness of our method by consistently extending it across different base algorithms, model architectures, and modalities.
    \item[] Guidelines:
    \begin{itemize}
        \item The answer \answerNA{} means that the paper does not include experiments.
        \item The authors should answer \answerYes{} if the results are accompanied by error bars, confidence intervals, or statistical significance tests, at least for the experiments that support the main claims of the paper.
        \item The factors of variability that the error bars are capturing should be clearly stated (for example, train/test split, initialization, random drawing of some parameter, or overall run with given experimental conditions).
        \item The method for calculating the error bars should be explained (closed form formula, call to a library function, bootstrap, etc.)
        \item The assumptions made should be given (e.g., Normally distributed errors).
        \item It should be clear whether the error bar is the standard deviation or the standard error of the mean.
        \item It is OK to report 1-sigma error bars, but one should state it. The authors should preferably report a 2-sigma error bar than state that they have a 96\% CI, if the hypothesis of Normality of errors is not verified.
        \item For asymmetric distributions, the authors should be careful not to show in tables or figures symmetric error bars that would yield results that are out of range (e.g., negative error rates).
        \item If error bars are reported in tables or plots, the authors should explain in the text how they were calculated and reference the corresponding figures or tables in the text.
    \end{itemize}

\item {\bf Experiments compute resources}
    \item[] Question: For each experiment, does the paper provide sufficient information on the computer resources (type of compute workers, memory, time of execution) needed to reproduce the experiments?
    \item[] Answer: \answerYes{} 
    \item[] Justification: Please refer to Sec.~\ref{sec:experimental_settings} for detailed information on the GPU configurations. Training time is also reported in each experimental table.
    \item[] Guidelines:
    \begin{itemize}
        \item The answer \answerNA{} means that the paper does not include experiments.
        \item The paper should indicate the type of compute workers CPU or GPU, internal cluster, or cloud provider, including relevant memory and storage.
        \item The paper should provide the amount of compute required for each of the individual experimental runs as well as estimate the total compute. 
        \item The paper should disclose whether the full research project required more compute than the experiments reported in the paper (e.g., preliminary or failed experiments that didn't make it into the paper). 
    \end{itemize}
    
\item {\bf Code of ethics}
    \item[] Question: Does the research conducted in the paper conform, in every respect, with the NeurIPS Code of Ethics \url{https://neurips.cc/public/EthicsGuidelines}?
    \item[] Answer: \answerYes{} 
    \item[] Justification: We have completed the verification and confirmed that all items comply with the requirements.

    \item[] Guidelines:
    \begin{itemize}
        \item The answer \answerNA{} means that the authors have not reviewed the NeurIPS Code of Ethics.
        \item If the authors answer \answerNo, they should explain the special circumstances that require a deviation from the Code of Ethics.
        \item The authors should make sure to preserve anonymity (e.g., if there is a special consideration due to laws or regulations in their jurisdiction).
    \end{itemize}

\item {\bf Broader impacts}
    \item[] Question: Does the paper discuss both potential positive societal impacts and negative societal impacts of the work performed?
    \item[] Answer: \answerNo{} 
    \item[] Justification: This paper focuses on training efficiency optimization for vision-language models. We do not foresee direct negative societal impacts, as any potential risks stem from the underlying LLMs rather than being introduced by our method. 
    \item[] Guidelines:
    \begin{itemize}
        \item The answer \answerNA{} means that there is no societal impact of the work performed.
        \item If the authors answer \answerNA{} or \answerNo, they should explain why their work has no societal impact or why the paper does not address societal impact.
        \item Examples of negative societal impacts include potential malicious or unintended uses (e.g., disinformation, generating fake profiles, surveillance), fairness considerations (e.g., deployment of technologies that could make decisions that unfairly impact specific groups), privacy considerations, and security considerations.
        \item The conference expects that many papers will be foundational research and not tied to particular applications, let alone deployments. However, if there is a direct path to any negative applications, the authors should point it out. For example, it is legitimate to point out that an improvement in the quality of generative models could be used to generate Deepfakes for disinformation. On the other hand, it is not needed to point out that a generic algorithm for optimizing neural networks could enable people to train models that generate Deepfakes faster.
        \item The authors should consider possible harms that could arise when the technology is being used as intended and functioning correctly, harms that could arise when the technology is being used as intended but gives incorrect results, and harms following from (intentional or unintentional) misuse of the technology.
        \item If there are negative societal impacts, the authors could also discuss possible mitigation strategies (e.g., gated release of models, providing defenses in addition to attacks, mechanisms for monitoring misuse, mechanisms to monitor how a system learns from feedback over time, improving the efficiency and accessibility of ML).
    \end{itemize}
    
\item {\bf Safeguards}
    \item[] Question: Does the paper describe safeguards that have been put in place for responsible release of data or models that have a high risk for misuse (e.g., pre-trained language models, image generators, or scraped datasets)?
    \item[] Answer: \answerNo{} 
    \item[] Justification:  This paper poses no such risks.
    \item[] Guidelines:
    \begin{itemize}
        \item The answer \answerNA{} means that the paper poses no such risks.
        \item Released models that have a high risk for misuse or dual-use should be released with necessary safeguards to allow for controlled use of the model, for example by requiring that users adhere to usage guidelines or restrictions to access the model or implementing safety filters. 
        \item Datasets that have been scraped from the Internet could pose safety risks. The authors should describe how they avoided releasing unsafe images.
        \item We recognize that providing effective safeguards is challenging, and many papers do not require this, but we encourage authors to take this into account and make a best faith effort.
    \end{itemize}

\item {\bf Licenses for existing assets}
    \item[] Question: Are the creators or original owners of assets (e.g., code, data, models), used in the paper, properly credited and are the license and terms of use explicitly mentioned and properly respected?
    \item[] Answer: \answerYes{} 
    \item[] Justification: All the assets used in this paper are properly credited and the license and terms
of use are explicitly mentioned and properly respected.
    \item[] Guidelines:
    \begin{itemize}
        \item The answer \answerNA{} means that the paper does not use existing assets.
        \item The authors should cite the original paper that produced the code package or dataset.
        \item The authors should state which version of the asset is used and, if possible, include a URL.
        \item The name of the license (e.g., CC-BY 4.0) should be included for each asset.
        \item For scraped data from a particular source (e.g., website), the copyright and terms of service of that source should be provided.
        \item If assets are released, the license, copyright information, and terms of use in the package should be provided. For popular datasets, \url{paperswithcode.com/datasets} has curated licenses for some datasets. Their licensing guide can help determine the license of a dataset.
        \item For existing datasets that are re-packaged, both the original license and the license of the derived asset (if it has changed) should be provided.
        \item If this information is not available online, the authors are encouraged to reach out to the asset's creators.
    \end{itemize}

\item {\bf New assets}
    \item[] Question: Are new assets introduced in the paper well documented and is the documentation provided alongside the assets?
    \item[] Answer: \answerYes{} 
    \item[] Justification: All the new assets introduced in this paper are well documented and the
documentation is provided alongside the assets.
    \item[] Guidelines:
    \begin{itemize}
        \item The answer \answerNA{} means that the paper does not release new assets.
        \item Researchers should communicate the details of the dataset\slash code\slash model as part of their submissions via structured templates. This includes details about training, license, limitations, etc. 
        \item The paper should discuss whether and how consent was obtained from people whose asset is used.
        \item At submission time, remember to anonymize your assets (if applicable). You can either create an anonymized URL or include an anonymized zip file.
    \end{itemize}

\item {\bf Crowdsourcing and research with human subjects}
    \item[] Question: For crowdsourcing experiments and research with human subjects, does the paper include the full text of instructions given to participants and screenshots, if applicable, as well as details about compensation (if any)? 
    \item[] Answer: \answerNo{} 
    \item[] Justification:  This paper does not involve crowdsourcing nor research with human subjects.
    \item[] Guidelines:
    \begin{itemize}
        \item The answer \answerNA{} means that the paper does not involve crowdsourcing nor research with human subjects.
        \item Including this information in the supplemental material is fine, but if the main contribution of the paper involves human subjects, then as much detail as possible should be included in the main paper. 
        \item According to the NeurIPS Code of Ethics, workers involved in data collection, curation, or other labor should be paid at least the minimum wage in the country of the data collector. 
    \end{itemize}

\item {\bf Institutional review board (IRB) approvals or equivalent for research with human subjects}
    \item[] Question: Does the paper describe potential risks incurred by study participants, whether such risks were disclosed to the subjects, and whether Institutional Review Board (IRB) approvals (or an equivalent approval/review based on the requirements of your country or institution) were obtained?
    \item[] Answer: \answerNo{} 
    \item[] Justification:  This paper does not involve crowdsourcing nor research with human subjects.
    \item[] Guidelines:
    \begin{itemize}
        \item The answer \answerNA{} means that the paper does not involve crowdsourcing nor research with human subjects.
        \item Depending on the country in which research is conducted, IRB approval (or equivalent) may be required for any human subjects research. If you obtained IRB approval, you should clearly state this in the paper. 
        \item We recognize that the procedures for this may vary significantly between institutions and locations, and we expect authors to adhere to the NeurIPS Code of Ethics and the guidelines for their institution. 
        \item For initial submissions, do not include any information that would break anonymity (if applicable), such as the institution conducting the review.
    \end{itemize}

\item {\bf Declaration of LLM usage}
    \item[] Question: Does the paper describe the usage of LLMs if it is an important, original, or non-standard component of the core methods in this research? Note that if the LLM is used only for writing, editing, or formatting purposes and does \emph{not} impact the core methodology, scientific rigor, or originality of the research, declaration is not required.
    \item[] Answer: \answerNo{} 
    \item[] Justification:  The core method development in this research does not involve LLMs as any
important, original, or non-standard components.
    \item[] Guidelines:
    \begin{itemize}
        \item The answer \answerNA{} means that the core method development in this research does not involve LLMs as any important, original, or non-standard components.
        \item Please refer to our LLM policy in the NeurIPS handbook for what should or should not be described.
    \end{itemize}

\end{enumerate}

\end{document}